\documentclass[10pt,twocolumn,letterpaper]{article}

\usepackage[pagenumbers]{cvpr} 

\definecolor{cvprblue}{rgb}{0.21,0.49,0.74}
\usepackage[pagebackref,breaklinks,colorlinks,allcolors=cvprblue]{hyperref}
\usepackage{layout}
\usepackage{hyperref}
\usepackage{multirow}
\usepackage{pifont}

\def\paperID{*****} 
\def\confName{CVPR}
\def\confYear{2026}

\title{The Role of World Models in Shaping Autonomous Driving: \\A Comprehensive Survey}

\author{
Sifan Tu$^{1}$, Xin Zhou$^{1}$, Dingkang Liang$^{1}$, Xingyu Jiang$^{1}$, Yumeng Zhang$^{2}$, Xiaofan Li$^{2}$, Xiang Bai$^{1}$\\
$^{1}$Huazhong University of Science and Technology, $^{2}$Baidu Inc.\\
{\tt\small \{sifantu, xzhou03, dkliang\}@hust.edu.cn}
}

\begin{document}
\maketitle
\begin{abstract}
The Driving World Model (DWM), which focuses on predicting scene evolution during the driving process, has emerged as a promising paradigm in the pursuit of autonomous driving (AD). DWMs enable AD systems to better perceive, understand, and interact with dynamic driving environments. In this survey, we provide a comprehensive overview of the latest progress in DWM. First, we review the DWM ecosystem, which is constructed using mainstream simulators, high-impact datasets, and various metrics that evaluate DWMs across multiple dimensions. We then categorize existing approaches based on the modalities of the predicted scenes, including video, point cloud, occupancy, latent feature, and traffic map, and summarize their specific applications in AD research. In addition, the performance of representative approaches across generating and driving tasks is presented. Finally, we discuss the potential limitations of current research and propose future directions. This survey provides valuable insights into the development and application of DWM, fostering its broader adoption in AD. The relevant papers are collected at \url{https://github.com/LMD0311/Awesome-World-Model}.
\end{abstract}
    
\section{Introduction}
\label{sec:1}
World Models are designed to forecast the future scene given historical observations, thereby internalizing the principles governing the external world~\cite{ha2018world}. This paradigm is regarded as a promising path toward human-level real-world cognition~\cite{lecun2022path}. In the domain of Autonomous Driving (AD), the inherent complexity of environmental dynamics and stringent operational-safety requirements pose significant challenges for traditional perception systems~\cite{hu2023planning}, while the scarcity of high-quality annotations limits their scalability. In this context, Driving World Models (DWMs) emerge as an instrumental methodology for addressing these hurdles. 

\begin{figure}[t]
\centering 
\includegraphics[width=0.99\linewidth]{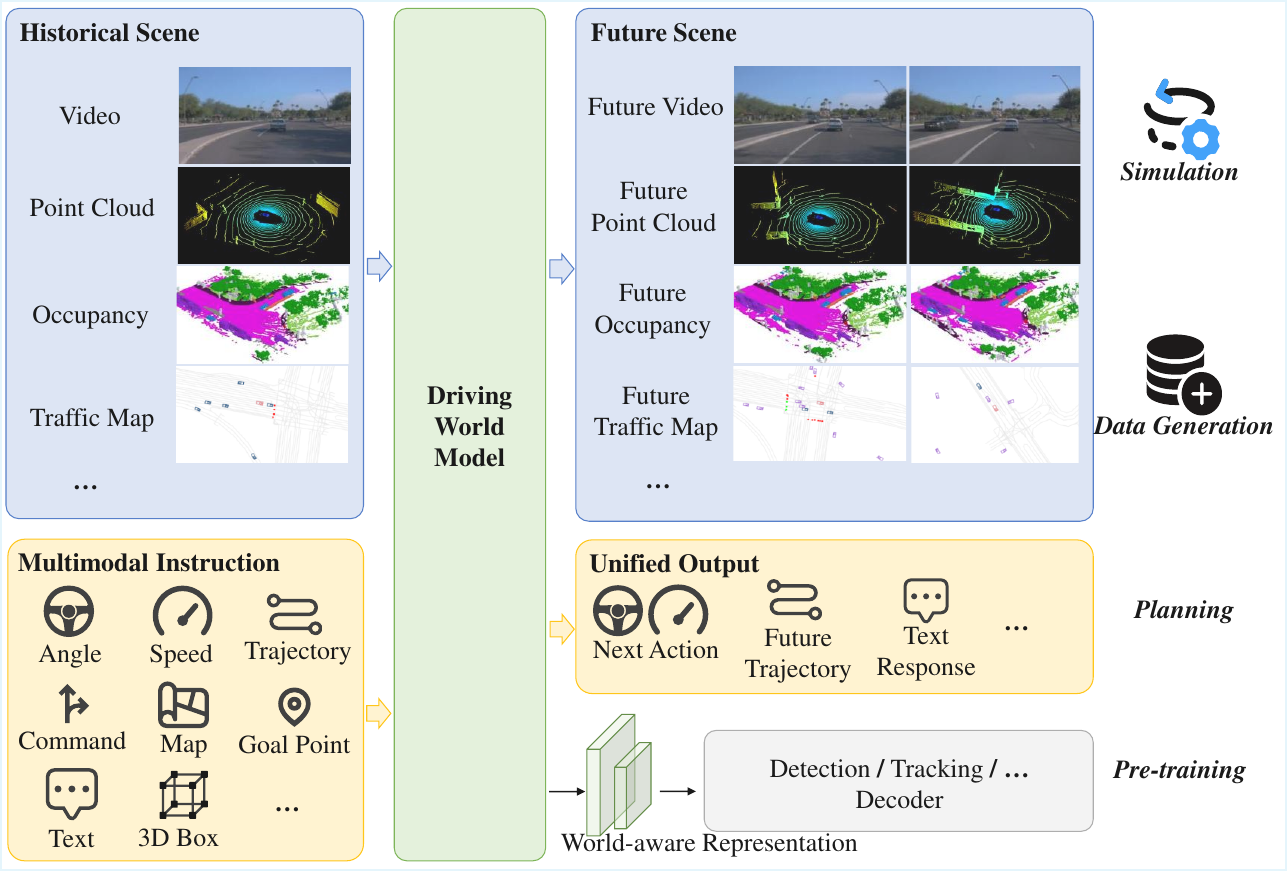} 
\caption{DWMs perceive historical observations and predict future scene evolutions, which adhere to multimodal conditions. This paradigm is applied to various domains of AD research, including simulation, data generation, driving, and pre-training.} 
\label{fig:dwm} 
\end{figure}

As illustrated in Fig.~\ref{fig:dwm}, the primary objective of DWMs is to forecast the temporal evolution of driving scenes conditioned on diverse instructions. This paradigm promotes category-agnostic perception, enabling a holistic understanding of complex environments. Furthermore, the DWM diagram facilitates self-supervised training, enabling the effective utilization of vast unlabeled data. Therefore, these methods are of great significance for the realization of reliable AD systems. 

Recently emerging research on DWMs encompass a diverse array of modalities within the AD research landscape, ranging from real-world representations, such as video~\cite{hu2023gaia,wang2023drivedreamer,gao2023magicdrive,ni2025maskgwm,chen2024unimlvg,russell2025gaia}, point clouds~\cite{yang2024visual,zhou2025hermes,zyrianov2024lidardm,liang2025lidarcrafter,zhou2025lagen} and occupancy~\cite{zheng2024occworld,wang2024occsora,wei2024occllama,yan2024renderworld,gu2024dome}, to latent features~\cite{karypidis2024dino,li2024enhancing,zheng2025world4drive,popov2024mitigating,li2024think2drive} and traffic maps~\cite{zhang2023trafficbots,sheng2025talk2traffic,jiang2024scenediffuser,tan2025scenediffuser++}. Each modality in DWM involves its own characteristics, resulting in unique developmental paths. For example, video-targeted DWMs~\cite{wang2024driving,gao2024vista,yang2024physical} focus on high-fidelity and multi-view-consistent generation, whereas occupancy-targeted DWMs~\cite{xu2025occ,zhang2024efficient,mei2025vision} place a greater emphasis on computational efficiency. On the other hand, the comprehensive impact of DWMs on AD research is also reflected in diverse application pathways. The controllable and realistic generation capability of DWMs enables driving scene simulation~\cite{yan2025drivingsphere,yang2025drivearena} and diverse data synthesis~\cite{wang2023drivedreamer,yang2024physical,zyrianov2024lidardm}, while the understanding and forecasting of dynamic scenes augment planners with robust perception~\cite{zheng2024occworld,wei2024occllama,li2024enhancing} and enriched contextual awareness~\cite{zhang2025future,zeng2025futuresightdrive}. In addition, DWM-based pre-training diagrams~\cite{yang2024visual,shi2025drivex,min2024driveworld,agro2024uno,huang2024neural} demonstrate a significant enhancement in the performance of various downstream tasks.

Diverse generative and functional capabilities foster a deep integration between DWMs and AD research. However, this variety simultaneously introduces considerable complexity, leading to a fragmented research landscape. Existing surveys~\cite{zhu2024sora,ding2024understanding,zeng2024world,fu2024exploring,guan2024world,jia2025progressive} tend to group DWMs within a single system, overlooking the unique challenges associated with specific implementations. In contrast, this survey provides a comprehensive review of DWM research and offers a detailed classification based on distinct prediction modalities and applications. By investigating these unique technical paths, this paper aims to provide an exhaustive reference for researchers entering the field and present insightful perspectives for future DWM research.

Our contributions can be summarized as follows:
\begin{itemize}
\item We provide a comprehensive review of the recent evolution of DWMs, categorizing methods by prediction modalities, along with a systematic analysis of their respective development paths.
\item We synthesize the integration of DWMs with AD research, elucidating their impact on simulation, data generation, driving performance, and training paradigms.
\item We conduct a critical analysis of current limitations and outline promising future directions, delivering valuable insights for the community and facilitating the further development of DWMs.
\end{itemize}

\begin{figure*}[t]
\centering 
\includegraphics[width=0.99\linewidth]{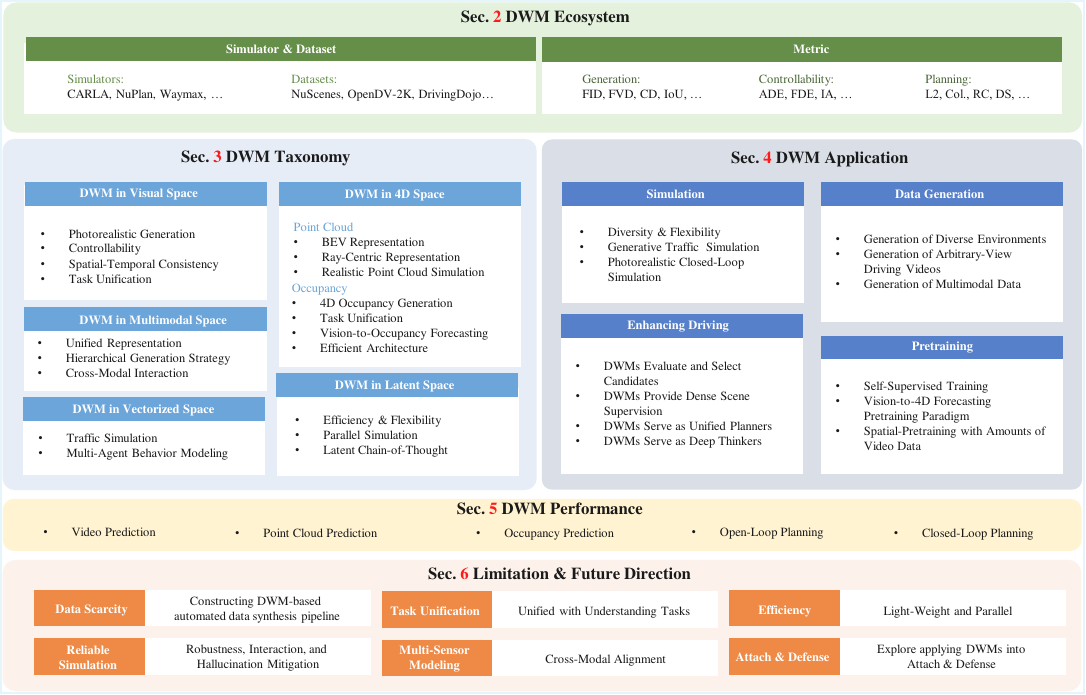} 
\caption{Main structure. Sec.~\ref{sec:2} introduces simulators, datasets, and metrics involved in DWM studies. Sec.~\ref{sec:3} and Sec.~\ref{sec:4} delineate the development of DWMs separately from prediction modalities and applications. Sec.~\ref{sec:5} presents the performance of DWMs across generation and planning tasks. Sec.~\ref{sec:6} outlines current limitations and future research directions. } 
\label{fig:main} 
\end{figure*}

The overall organization of this survey is illustrated in Figure~\ref{fig:main}. Sec.~\ref{sec:2} introduces mainstream simulators, high-impact datasets, and multi-dimensional metrics of current DWMs. Sec.~\ref{sec:3} and Sec.~\ref{sec:4} discuss existing approaches from the perspectives of prediction modalities and applications, respectively. Sec.~\ref{sec:5} presents the performance of DWMs on primary tasks, including generation tasks of various modalities and driving planning. Finally, Sec.~\ref{sec:6} summarizes the current limitations and outlines future research directions.

\section{DWM Ecosystem}\label{sec:2}
Simulators, datasets, and metrics form the ecosystem of DWM search, serving as the foundation for training and evaluation. Early DWMs~\cite{wang2023drivedreamer,wen2024panacea,li2024drivingdiffusion,pan2022iso} are trained and evaluated on traditional AD datasets and simulators, with their capabilities measured by generation and driving planning metrics. Following the rapid development of this field, an ecosystem specifically for DWMs is gradually taking shape. In this section, we conduct a systematic review of existing simulators, datasets, and metrics to clarify how these factors work together to shape the current research landscape of DWMs.

\begin{table*}[t]
    \scriptsize
    \centering
    \caption{Mainstream simulators. Full-stack simulation encompasses the simulation of both driving behaviors and observed scenarios, whereas behavior simulation refers to the simulation of driving behaviors alone.}
    \setlength{\tabcolsep}{1.7mm}
    \begin{tabular}{lccl}
    \toprule
\textbf{Platform} & \textbf{Year} & \textbf{Domain} & \textbf{Description} \\
\midrule
CARLA~\cite{dosovitskiy2017carla} & 2016 & Full-Stack Simulation & Providing multimodal scene data, along with comprehensive annotations. \\
Nvidia Drive Sim~\cite{nvidiasim2024} & 2020 & Full-Stack Simulation &  Simulation and synthetic data generation pipelines powered by NVIDIA’s advanced models, tools, and libraries.\\
MetaDrive~\cite{li2022metadrive} & 2021 & Behavior Simulation & Featuring scene composability and scalability to accommodate diverse reinforcement learning training tasks. \\
NuPlan~\cite{caesar2021nuplan} & 2021 & Behavior Simulation & Providing a large-scale driving dataset and an open-source traffic simulator.\\
Waymax~\cite{gulino2023waymax} & 2023 & Behavior Simulation & Multi-agent behavior simulator based on the Waymo Open Dataset~\cite{peebles2023scalable}. \\
LimSim~\cite{wen2023limsim} & 2023 & Behavior Simulation & Long-term interactive traffic simulator. \\
\bottomrule  
    \end{tabular}
    \label{tab:simulator}
\end{table*}

\begin{table}[ht]
    \scriptsize
    \centering
    \caption{High-impact datasets. \# Scene reflects the dataset diversity, while Duration indicates the dataset scale.}
    \setlength{\tabcolsep}{0.5mm}
    \begin{tabular}{lccc}
    
    \toprule
    Dataset & \# Scene & Duration (h) & Example\\
    \midrule
    KITTI~\cite{Geiger2013IJRR} & 22 & 1.5 & \cite{zhang2023learning} \\
    
    Waymo~\cite{sun2020scalability} & 1k & 6.4 & \cite{guo2025dist} \\
    Arg 2 LiDAR~\cite{Argoverse2} & 20k & 167 & \cite{guo2025dist} \\
    nuScenes~\cite{caesar2020nuscenes} & 1k & 5.5 & \cite{gu2024dome}~\cite{yang2024visual}~\cite{zhou2025hermes}~\cite{xu2025occ}~\cite{yan2024renderworld}~\cite{zuo2024gaussianworld} \\
    nuPlan~\cite{caesar2021nuplan} & - & 1200 & \cite{hu2024drivingworld} \\
    NAVSIM~\cite{dauner2024navsim} & 115k & - & \cite{chen2024drivinggpt,zheng2025world4drive,zhang2025epona,li2024enhancing,li2025end} \\
    OpenScene~\cite{openscene2023} & - & 120 & \cite{zhang2024efficient}~\cite{min2024driveworld} \\
    Occ3D-nuScenes~\cite{tian2023occ3d} & 1k & 5.5 & \cite{zheng2024occworld}~\cite{shi2025come}~\cite{bian2024dynamiccity}~\cite{wang2024occsora}~\cite{xu2025occ} \\
    OpenDV-YouTube~\cite{yang2024genad} & - & 1700 & \cite{gao2024vista}~\cite{yang2024genad} \\
    OpenDV-2K~\cite{yang2024genad} & - & 2059 & \cite{ni2025maskgwm} \\
    Cityscapes~\cite{cordts2016cityscapes} & 25k & 12.5 & \cite{karypidis2025advancing} \\
    OpenOccupancy~\cite{wang2023openoccupancy} & 1.1k & 5.5 & \cite{yang2024driving}~\cite{wei2024occllama}~\cite{li2025uniscene}~\cite{chenoccprophet} \\
    DIVA\cite{zhou2024simgen} & 135 & 121 & \cite{zhou2024simgen} \\
    DrivingDojo~\cite{wang2024drivingdojo} & 18k & 150 & \cite{hassan2024gem} \\
    ACT-Bench~\cite{arai2024act} & 2.3k & - & - \\
    Cam4DOcc~\cite{ma2024cam4docc} & 24k & - & \cite{yang2024driving}\\
    UniOcc~\cite{wang2025uniocc} & 2.2k & 14.2 & - \\
    Bench2Drive~\cite{jia2024bench2drive} & 10k & - & \cite{li2024think2drive}~\cite{yang2025raw2drive} \\      
    WorldSimBench~\cite{qin2024worldsimbench} & 36k & - & - \\
    \bottomrule
    \end{tabular}
    \label{tab:dataset}
\end{table}

\begin{table*}[!ht]
    \scriptsize
    \centering
    \setlength{\tabcolsep}{1.9mm}
    \caption{Multi-dimensional metrics. $\uparrow$ indicates better performance with higher values, while $\downarrow$ indicates better performance with lower values.}
    \begin{tabular}{lccl}
    \toprule
    Metric & $\downarrow$ or $\uparrow$ & Task & Description \\
    \midrule
    Fréchet Inception Distance (FID) & $\downarrow$ & Image / Video Generation & Feature distribution distance between the predicted and GT images. \\ 
    Fréchet Video Distance (FVD) & $\downarrow$ & Video Generation & Feature distribution distance between the predicted and GT videos.\\
    Chamfer Distance (CD) & $\downarrow$ & Point Cloud Generation & Average bidirectional nearest neighbor distance between the predicted and GT point clouds.\\
    Displacement Error (L2) & $\downarrow$ & Open-loop Planning &  L2 distance between the predicted and GT trajectory.\\
    Collision Rate (Col.) & $\downarrow$ & Open-loop Planning & Rate of collisions that occur with other objects when following the predicted trajectory.\\
    \midrule
    Average Displacement Error (ADE) & $\uparrow$ & Video Generation & Average point-wise error between actual and predicted trajectories. \\
    Final Displacement Error (FDE) & $\uparrow$ & Video Generation & Error between the actual and predicted positions at the final time step.\\
    Aesthetics (AE) & $\uparrow$ & Video Generation & Visual quality of generated video.\\
    Instruction Alignment (IA) & $\uparrow$ & Video Generation & Whether the video aligns with the provided instruction.\\
    Perspectivity (PV) & $\uparrow$ & Video Generation & Whether the video has spatial realism.\\
    Trajectory (TJ) & $\uparrow$ & Video Generation & Whether the movement and the trajectory of elements in the video are logical.\\
    Key Element (KE) & $\uparrow$ & Video Generation & The generated quality of embodied elements.\\
    Safety (SF) & $\uparrow$ & Video Generation & Whether the behavior of the vehicles complies with traffic rules.\\
    Intersection over Union (IoU) & $\uparrow$ & Occupancy Generation & Ratio of the intersection and union between the predicted and GT occupancy.\\
    Mean IoU (mIoU) & $\uparrow$ & Occupancy Generation & Average of IoU values across all categories.\\
    Route Completion (RC) & $\uparrow$ & Closed-loop Planning & Percentage of the route completed by ego. \\
    Infraction Score (IS) & $\uparrow$ & Closed-loop Planning & Initialized at 1.0 and decreases by a penalty coefficient for each Infraction.\\
    Driving Score (DS) & $\uparrow$ & Closed-loop Planning & Product of RC and IS for each sample.\\
    Success Rate (SR) & $\uparrow$ & Closed-loop Planning & Proportion of routes completed successfully on time and without traffic violations. \\
    Efficient (Effi.) & $\uparrow$ & Closed-loop Planning & Mean of ego speed versus average surrounding traffic speed ratio per sampling point.\\
    Comfortness (Comf.) & $\uparrow$ & Closed-loop Planning & Ratio of smooth trajectory segments to the total number of segments. \\
    \bottomrule   
    \end{tabular}
    \label{tab:metric}
\end{table*}

\subsection{Simulator \& Dataset}
Simulators and datasets serve as the fundamental basis of DWM research. Specifically, simulators enable the acquisition of interactive data within virtual scenarios, while datasets facilitate the training and evaluation of frameworks with real-world data, thereby enhancing their generalization capabilities in the physical world.

\textbf{Simulators.} Simulators enable large-scale closed-loop training and evaluation, serving as the foundation of the DWM ecosystem. Tab.~\ref{tab:simulator} provides a comprehensive overview of the simulators widely utilized in DWM research. We categorize these platforms based on their functional scope. Full-stack simulators~\cite{dosovitskiy2017carla,nvidiasim2024} support the generation of both agent behaviors and sensory observations, while Behavior simulators~\cite{gulino2023waymax,caesar2021nuplan,li2022metadrive} focus on agent dynamics and logic. Notably, behavior simulators can be integrated with scene rendering models to construct full-stack simulations~\cite{yan2025drivingsphere,yang2025drivearena}, demonstrating a feasible path towards photorealistic closed-loop simulation.

\textbf{Datasets.} Tab.~\ref{tab:dataset} presents the primary datasets in current DWM research. Traditional AD datasets such as KITTI~\cite{Geiger2013IJRR}, Waymo Open Dataset~\cite{sun2020scalability}, and nuScenes~\cite{caesar2020nuscenes}, characterized by comprehensive data modalities and meticulous annotations, serve as a pivotal driving force for DWM research. However, the majority of these driving datasets require field acquisition and considerable cost, which restricts the scale and diversity of the dataset and gradually becomes a bottleneck amidst the rapid advancement of self-supervised DWMs~\cite{yang2024visual,li2025semi,min2024driveworld}.

While datasets like OpenDV-YouTube~\cite{yang2024genad} attempt to address this by harvesting large-scale data from the internet, they fail to adequately compensate for the extreme scarcity of long-tail scenarios and the lack of multi-modal data. Recent studies try to construct datasets rich in complex driving behaviors, aiming to address the urgent demand within DWM research for large-scale, diverse data. For instance, DrivingDojo~\cite{wang2024drivingdojo} conducts field collection and meticulous filtering of large-scale driving videos to establish a diverse dataset covering ego-actions, multi-agent interplay, and open-world knowledge.

With the continuous progress of DWM, new characteristics and requirements gradually emerge, such as vision-to-4D prediction and controllable action simulation. In parallel, New datasets and benchmarks, including Cam4DOcc~\cite{ma2024cam4docc}, ACT-Bench~\cite{arai2024act}, and WorldSimBench~\cite{qin2024worldsimbench}, are also introduced to satisfy the evaluation needs of existing approaches and guide the direction of subsequent studies.

\subsection{Metric}
Tab.~\ref{tab:metric} presents metrics previously used in DWM research. In most DWM research, the objectives are largely indistinguishable from those of standard generative tasks. Researchers prioritized the global similarity between data synthesized by the DWM and ground-truth (GT) data. A model exhibiting higher generation fidelity (or lower error) is deemed to have more effectively acquired the world knowledge inherent in driving scenarios. Consequently, during this stage, the capabilities of DWMs are predominantly evaluated using generative metrics. Specifically, video generation tasks utilize FID and FVD to quantify the divergence between the generated and GT data. Point cloud generation tasks employ CD to measure the average bidirectional nearest neighbor distance between the generated and GT point sets. Meanwhile, occupancy generation tasks relie on IoU and mIoU to assess the degree of volumetric overlap between the generated and GT occupancy grids. 

However, generative metrics often place disproportionate emphasis on low-level scene details, thereby obscuring the evaluation of the model's ability to internalize underlying environmental dynamics. 
Therefore, several benchmarks introduce additional metrics to conduct a multi-dimensional assessment of DWM generation quality. For instance, ACT-BENCH~\cite{arai2024act} employs ADE and FDE to determine if the generated videos follow behavioral commands, whereas WorldSimBench~\cite{qin2024worldsimbench} utilizes various metrics (e.g., AE, IA, PV, TJ, KE, and SF) to evaluate different video properties relevant to AD, such as video quality, controllability, and path validity.

Finally, DWMs can be indirectly assessed by evaluating the performance of the frameworks into which they are integrated. Comparing planning performance is a common paradigm for gauging the impact of DWMs on the overall framework. For example, metrics such as L2 and Col. in open-loop evaluation, and DS and SR in closed-loop evaluation, are all applicable in related studies. This broad applicability underscores the flexibility and generalization capabilities of DWMs. Furthermore, evaluation on various downstream tasks can also serve to validate the quality of DWM-generated data~\cite{yang2024physical,wang2023drivedreamer,wen2024panacea} while also reflecting the generalizability of pre-trained DWMs~\cite{yang2024visual,min2024driveworld,wang2025uniocc,du2025sparseworld,shi2025drivex}.

\section{DWM Taxonomy}
\label{sec:3} 
DWMs involve driving scenarios of various modalities, such as video, point clouds, occupancy, latent features, and traffic map. Each modality offers distinct advantages as well as corresponding challenges. This section reviews the latest research progress in DWMs from the perspective of prediction modalities and summarizes the development pathways of these models for each modality.

\subsection{DWM in Visual Space}
Video data offers distinct advantages, including low acquisition costs, high resolution, and rich semantic and textural information, making video-targeted DWM a pivotal research direction. However, its reliance on pixel distributions to indirectly represent observed 3D space often compromises localization accuracy. Furthermore, the susceptibility of camera sensors to environmental variables, such as adverse weather and fluctuating illumination, limits their robustness in complex driving scenarios. The core objective of video-targeted DWMs is to learn spatiotemporal dynamics and physical laws from visual inputs via driving video prediction. Early approaches~\cite{pan2022iso,hu2022model,gao2024enhance} are constrained by the generative techniques available at the time, and are therefore trained and evaluated within simulated scenarios. In recent years, the application of generative technologies~\cite{rombach2022high,li2024drivingdiffusion,peebles2023scalable} has propelled video-targeted DWMs toward predicting complex real-world scenes.

Diffusion-based DWMs~\cite{wang2023drivedreamer,wen2024panacea,li2024drivingdiffusion,gao2023magicdrive} demonstrate superior generation capabilities characterized by fine-grained controllability. Based on several reference frames, they can synthesize photorealistic long-term driving videos of diverse environments and views that adhere to multimodal instructions, including layouts, 3D boxes, and actions. This paradigm aims to achieve the goal of generating high-fidelity, flexibly controllable, and spatiotemporally consistent driving videos.

Conditions provide Diffusion-based DWMs with motion and scene priors. Through improved alignment with generative modules, these conditions enable flexible controllability and high-quality generation. For example, SimGen~\cite{zhou2024simgen} and Dreamland~\cite{mo2025dreamland} transform diversity-constrained scene conditions into forms that better conform to real-world distributions, while DriveDreamer-2~\cite{zhao2024drivedreamer} converts textual instructions into structured conditions, enabling accessible control of complex driving scenes. Additionally, GAIA-2~\cite{russell2025gaia} and Vista~\cite{gao2024vista} establish multi-level scene control mechanisms via diverse conditions, while UniMLVG~\cite{chen2024unimlvg} fuses text instructions with 3D conditions, thereby supporting diverse generation paradigms. To enhance realism, DrivePhysica~\cite{yang2024physical} introduces physical priors, such as global coordinates, as control conditions. Furthermore, predicting spatial-aware representations beforehand to provide 4D prior~\cite{lu2024wovogen,ji2025cogen,chen2025geodrive} also significantly improves spatial plausibility and spatial consistency.

On the other hand, advancements in model architectures and training strategies significantly advance the spatiotemporal consistency of DWMs. Specifically, Drive-WM~\cite{wang2024driving} introduces a masked view prediction task, whereas MagicDrive-V2~\cite{gao2025magicdrive} and DiVE~\cite{jiang2025dive} design view interaction modules. These architectures model multi-view correlations, significantly improving the consistency across multiple views in the predicted videos. For long-horizon generation, multi-temporal granularity prediction mechanisms are introduced to enhance stability and improve performance. For instance, InfinitiDrive~\cite{guo2024infinitydrive} achieves efficient spatiotemporal modeling through parallel branches with distinct sampling rates, while MiLA~\cite{wang2025mila} and LongDWM~\cite{wang2025longdwm} establish a coarse-to-fine progressive paradigm. In addition, Epona~\cite{zhang2025epona} employs multi-round propagation conditioned on generated frames, enhancing adaptation to errors during long-horizon rollouts and mitigating degradation phenomena.

Another pathway, Autoregressive-Transformer-based DWMs, explores handling multiple tasks within a unified sequence. These approaches aim to facilitate the integration of world knowledge learned from generation and driving tasks, thereby enhancing overall performance. In detail, GAIA-1~\cite{hu2023gaia} demonstrates that DWMs can learn scene distributions from large-scale data, while DrivingGPT~\cite{chen2024drivinggpt} and DrivingWorld~\cite{hu2024drivingworld} refine the unification of scene and action representations. Furthermore, Doe-1~\cite{zheng2024doe} shifts focus towards the closed-loop unified modeling of perception, prediction, and planning tasks.

In conclusion, the advancement of vision-targeted DWMs presents distinct challenges in modeling rapid scene dynamics and ensuring multi-view spatial consistency. With the development of generative technologies, research paradigms evolve from simulator-based training to leveraging robust generative baselines. Diffusion-based approaches successfully enhance fidelity and controllability, while Autoregressive Transformer architectures pioneer the unified modeling of diverse driving tasks within a single sequence. However, achieving robust long-horizon video generation without error accumulation and effectively embedding rigorous physical laws to support reliable closed-loop decision-making in open-world scenarios remains a pivotal area yet to be fully realized.

\subsection{DWM in 4D Space}
4D data refers to data modalities that comprise 3D spatial and 1D temporal information. These data inherently preserve structural consistency, detailed geometric information, and precise spatiotemporal dynamics, and are extensively adopted in AD research~\cite{zhou2024dynamic,liang2024pointmamba,tong2023scene}, emerging as a pivotal focus in the development of DWMs. Here, we introduce primary 4D data modalities (i.e., point cloud sequence and occupancy sequence), and review the corresponding DWM research within their respective domains.

\textbf{Point Cloud.} LiDAR sensors acquire point clouds through laser reflection, ensuring robustness against environmental interference such as low illumination and smoke. In contrast to 2D pixels, point clouds provide precise depth measurements and explicit geometric structure, which are essential for achieving high-fidelity 4D generation. However, the intrinsic irregularities of this data, particularly sparsity and lack of ordered topology, present substantial difficulties for modeling in long-horizon and highly dynamic scenarios.

The core challenge in point cloud forecasting lies in accurately predicting the geometric evolution within continuous 3D space. Direct regression of 3D point coordinates is not only computationally prohibitive but also highly sensitive to sensor-specific characteristics. To address this, Bird's-Eye-View (BEV) representations serve as a promised dimensionality reduction solution, projecting sparse, unordered 3D points onto structured grids to lower complexity while preserving spatial topology. Copilot4D~\cite{zhang2023learning} encodes unstructured point cloud data into discrete BEV token sequences, demonstrating a scalable method for point-cloud-targeted DWMs. Additionally, the viewpoint retention of BEV facilitates modality transformation. For instance, ViDAR~\cite{yang2024visual} constructs a vision-to-point-cloud forecasting framework via BEV, achieving broad generalization across various downstream tasks. Taking a step further, HERMES~\cite{zhou2025hermes} unifies point cloud forecasting, driving planning, and scene question answering, while integrating textual world knowledge into the prediction of BEV features. As a result, it enables a synergistic interplay between understanding and generation. 

The ray-scanning mechanism of LiDARs inevitably results in a significant reduction in point density at long ranges, whereas the BEV representation projects features onto uniform grids. Due to the misalignment between grid uniformity and the inherent divergence of sensor beams, this transformation often results in the loss of detail for distant objects. To address this, LiSTAR~\cite{liu2025listar} introduces a ray-centric representation paradigm that aligns with the sensor's native scanning geometry, mitigating morphological distortion and enabling the generation of 4D point cloud scenes that are both physically realistic and precisely controllable. Similarly, LaGen~\cite{zhou2025lagen} processes point clouds into more compact range images via spherical projection, facilitating spatiotemporally consistent long-range point cloud prediction.

Point cloud DWMs extend beyond merely comprehending scene dynamics and forecasting future evolution, while facilitating the synthesis of 4D driving scenes characterized by physical plausibility, temporal consistency, and fine-grained controllability. Specifically, LidarDM~\cite{zyrianov2024lidardm} synthesizes background and object point clouds, selecting suitable paths to compose them into dynamic scenes. While capable of autonomously generating point cloud digital assets, it remains dependent on real trajectory data. In contrast, LiDARCrafter~\cite{liang2025lidarcrafter} establishes a multi-stage pipeline spanning from trajectory and object generation to dynamic scene construction, thereby enabling dynamic point cloud synthesis and fine-grained editing driven solely by language instructions. In addition, U4D~\cite{xu2025u4d} employs a hard-to-easy two-stage generation strategy to improve reconstruction in uncertain regions, such as fine-grained structures and distant areas, effectively reinforcing the overall reliability of 4D point cloud synthesis.

In the research of point cloud forecasting, the inherent sparsity and topological disorder of point cloud data pose significant challenges to efficient spatiotemporal modeling. To address this, the adoption of representations such as BEV, ray-based representations, and range images successfully strikes a balance between computational scalability and topological structure preservation. In the realm of 4D scene generation, research paradigms gradually transition from individual digital asset generation to full-scene generation, capable of fine-grained editing, which further enhances reliability in complex regions. However, modeling rare events or high-complexity environments, alongside the development of 4D generation paradigms that simultaneously balance accuracy and efficiency, remains a pivotal area yet to be explored. Transforming point clouds from the spatial domain~\cite{liang2024pointgst} to the spectral domain represents a promising solution.

\textbf{Occupancy.} Occupancy data represents 3D scenes through voxelization, which determines the existence of objects within discrete volumetric units. By assigning semantic labels to each occupied voxel, this representation constitutes semantic occupancy. A critical advantage of occupancy lies in its emphasis on geometric structure over predefined categorization. This property enables the explicit modeling of general obstacles independent of predefined taxonomies, a capability that facilitates collision avoidance for open-vocabulary objects. Its capacity for geometrically consistent volumetric modeling and structured spatial encoding makes occupancy exceptionally suitable for modeling 4D scene evolution, leading to its widespread adoption.

Different objectives correspond to different generation paradigms. Diffusion-based paradigms~\cite{wang2024occsora,gu2024dome,shi2025come,bian2024dynamiccity} typically encode occupancy into noisy continuous features. Under the guidance of multimodal conditions, these methods utilize core frameworks, such as Diffusion Transformers~\cite{peebles2023scalable}, to forecast future states, thereby achieving high-quality generation characterized by realism, controllability, and temporal consistency. In contrast, Autoregressive-Transformer-based paradigms~\cite{zheng2024occworld,wei2024occllama,yan2024renderworld,xu2025occ,zhang2024efficient} encode occupancy scenes into latent space and serialize them with data of other modalities (e.g., trajectories and text) into a unified sequence. By leveraging the next-token-prediction mechanism, these approaches unify diverse tasks ranging from scene prediction and behavior planning to Occupancy Question Answering. Additionally, OccTENS~\cite{jin2025occtens} applies the next-scale-prediction paradigm~\cite{tian2024visual} to 4D occupancy prediction, exploring a novel autoregressive generative paradigm.

Despite the outstanding performance of these occupancy-targeted approaches, they suffer from severe performance degradation when relying solely on visual inputs, a limitation that significantly impedes their deployment in end-to-end AD systems. To address this challenge, RenderWorld~\cite{yan2024renderworld} devises a self-supervised image-to-occupancy translation module, while Cam4DOcc~\cite{ma2024cam4docc} explores end-to-end vision-to-4D occupancy forecasting. Furthermore, Drive-OccWorld~\cite{yang2024driving} constructs a vision-to-occupancy DWM adapted to diverse control conditions, integrating it with a planner to select the optimal trajectory. Although challenging, the paradigm of vision-to-4D occupancy forecasting is demonstrated to enhance DWMs to capture precise spatiotemporal dynamics from visual inputs. Accordingly, it is leveraged for 4D pre-training~\cite{min2023uniworld,min2024driveworld} and demonstrates effective generalization across diverse downstream tasks. Additionally, PreWorld~\cite{li2025semi} utilizes video data for 4D pre-training, enhancing semantic supervision while showcasing the potential for scaling to significantly larger datasets.
 
Real-world scenarios are dominated by extensive static backgrounds, which not only hold minimal relevance to driving behaviors but also interfere with the prediction of dynamic objects. Hence, decoupling strategies are widely adopted in occupancy-targeted DWMs. For example, Occ-LLM~\cite{xu2025occ} encodes static backgrounds and dynamic objects separately, thereby shifting the computational focus towards the evolution of dynamic trajectories. Similarly, DFIT-OccWorld~\cite{zhang2024efficient} and GaussianWorld~\cite{zuo2024gaussianworld} predict exclusively dynamic objects and newly observed regions, while the state changes of the static scene are directly mapped via pose transformations. These object-background decoupling approaches are simple and effective, but they remain constrained by predefined semantic labels, making it difficult to accommodate the complex dynamics in real-world driving situations. Rather than conducting object-specific predictions, implicit decoupling strategies~\cite{mei2025vision,xu2025temporal} estimate the global state transitions induced by the coupling of ego-motion and object dynamics. This approach not only enhances modeling efficiency but also prevents the omission of details. Moreover, other methods~\cite{liao2025i2,jin2025occtens} simultaneously model inter-frame dynamics and intra-frame structure, further bolstering spatiotemporal modeling capabilities and controllability.

On the other hand, constructing rational and flexible 4D scene representations constitutes another pivotal strategy for enhancing model efficiency. BEV representations are widely adopted~\cite{wei2024occllama,gu2024dome,shi2025come,yang2024driving,zhang2024efficient,li2025semi} due to their effective compression of spatial complexity. Furthermore, OccProphet~\cite{chenoccprophet} augments BEV with height and global features and preserves 3D structural information, whereas DTT~\cite{xu2025temporal} and DynamicCity~\cite{bian2024dynamiccity} employ multi-plane projection representations. More recently, sparse representations~\cite{du2025sparseworld,dang2025sparseworld} have garnered increasing attention. This paradigm models state transitions by leveraging a set of queries that adaptively traverse and interact with the original scene, exhibiting superior flexibility and holding significant potential for scaling to large-scale 4D scenes. In addition, UNO~\cite{agro2024uno} and DIO~\cite{diehl2025dio} leverage a novel 4D occupancy form, continuous occupancy fields, instead of occupancy grids. They sampled both occupied and non-occupied points, ensuring a balanced data distribution while simultaneously enhancing computational efficiency and structural granularity.

In conclusion, the evolution of occupancy-targeted DWMs is driven by the dual pursuit of robust, realistic 4D prediction and multi-task unification. While the occupancy paradigm enables precise representation, it simultaneously introduces challenges regarding significantly increased computational complexity and data acquisition costs. To address this, existing research has adopted strategies such as decoupled prediction and sparse representations, achieving remarkable results in processing efficiency. Nevertheless, despite attempts to leverage video data, the scarcity of direct large-scale 4D data persists. Addressing this data gap remains a significant objective for future work.

\subsection{DWM in Multimodal Space}
\label{sec:multimodal}
Visual data provides high-resolution texture and rich semantics, whereas 3D data captures precise spatiotemporal dynamics and inter-object relationships. Although these modalities prioritize different attributes, they describe the same underlying physical reality, permitting mutual translation and synergistic integration. The implementation of this concept has emerged within single-modal generative DWMs. For instance, video-targeted DWMs~\cite{zhang2025cvd,lu2024wovogen,ji2025cogen} integrate spatial representations into their pipelines to enhance layout accuracy and spatial consistency, while 4D-data-targeted DWMs leverage visual data to facilitate training~\cite{ost2025lsd,zhang2024efficient} or to forecast future 4D scenes conditioned on visual observations~\cite{yang2024visual,zhou2025hermes,min2023uniworld,min2024driveworld}. However, these methodologies remain focused on single data modalities, rendering them inadequate for the multi-sensor, multimodal operational architectures of modern AD systems. To address this gap, Multimodal generative DWMs are proposed, which entail forecasting future scenes across diverse data modalities, including video, occupancy, point cloud, depth, and even traffic maps. While this paradigm confronts the significant challenge of inherent data heterogeneity, it simultaneously offers the potential to leverage the intrinsic correlations that naturally exist between modalities of the same scene.

A fundamental challenge in multimodal DWMs is constructing a unified representation that can simultaneously capture both spatiotemporal dynamics and semantic information. Since depth maps and monocular videos share the same views, effective fusion can be readily achieved via direct concatenation~\cite{hassan2024gem,karypidis2025advancing,wu2025moviedrive}. In contrast, fusing multi-view videos with 4D scene data, such as occupancy and point clouds, presents a substantially greater challenge due to the significant heterogeneity in their representational modalities. MUVO~\cite{bogdoll2023muvo} addresses this by employing sequence concatenation and self-attention mechanisms to fuse multimodal information, while UMGen~\cite{wu2025generating} unifies the modeling of images and traffic maps within an autoregressive framework and fuses them with cross-attention blocks. On the other hand, NeMo~\cite{huang2024neural} constructs volumetric representations for detailed object modeling. More recently, capitalizing on the ability of BEV to preserve geometric spatial correspondences across diverse views, mainstream methodologies~\cite{zhang2024bevworld,shi2025drivex,tang2025omnigen} construct the unified multimodal data in BEV representations, thereby facilitating the comprehensive fusion and mutual enhancement of multimodal predictions. Furthermore, GaussianDWM~\cite{deng2025gaussiandwm} utilizes language guidance to sample from dense Gaussian Splatting~\cite{zhen20243d}, thereby establishing a joint image-depth generation framework that integrates world knowledge with visual features.

Despite the scalability and flexibility of the unified representations, the single-step fitting paradigm often struggles to characterize the complex data distribution characteristics inherent in real-world driving scenarios. To address this, some approaches~\cite{li2025uniscene,li2025scaling,lu2025infinicube} introduce 4D semantic occupancy as the intermediate prediction targets to model these complex distributions, achieving large-scale and physics-aware data synthesis through hierarchical generation strategies. 

In addition, the interaction mechanisms between different modal representations constitute another promising direction for exploration. Unlike the conventional practice of interacting via a unified representation, UniFuture~\cite{liang2025seeing} derives depth representations from visual features and establishes adaptive interaction through feedback branches, thereby achieving a mutual reinforcement between perception and generation. Similarly, HoloDrive~\cite{wu2024holodrive} constructs separate generators for video and point clouds and fuses multimodal information via dual projecting. This mechanism of adaptive interaction with decoupled representations demonstrates the potential for better accommodating modal biases, yielding superior quality in the generation process.

In summary, multimodal DWMs represent a pivotal breakthrough that bridges the semantic richness of visual data with the geometric precision of 3D representations, effectively addressing the limitations inherent in the heterogeneity of single-modal frameworks. However, most methods remain confined to specific modality combinations. Future foundational multimodal DWMs are expected to learn modality-agnostic world dynamics from observational data, thereby supporting the generation and transformation of arbitrary modalities.

\subsection{DWM in Latent Space}
\label{sec:latent}
Distinct from paradigms that strive for high-fidelity prediction of real-world scenes, latent DWMs focus on predicting latent representations fitting task-oriented AD frameworks. This design affords greater architectural flexibility, thereby effectively empowering AD research. 

Latent DWMs can be seamlessly integrated into existing frameworks, enhancing their performance. For instance, LAW~\cite{li2024enhancing} and World4Drive~\cite{zheng2025world4drive} predict future scene features conditioned on planned trajectories, enabling joint self-supervised training of scene perception and motion planning. AdaWM~\cite{wang2025adawm} adaptively fine-tunes the planner by evaluating the discrepancy between the latent DWM's predictions and the true next states, significantly improving adaptability to novel scenes. Moreover, LatentDriver~\cite{xiao2024learning} injects latent DWM outputs into the planner to synthesize physically plausible stochastic driving behaviors. Furthermore, DINO-Foresight~\cite{karypidis2024dino} and AD-L-JEPA~\cite{zhu2025ad} concentrate on high-level semantic representation learning rather than low-level detail noise. This strategy not only mitigates the interference of scene noise on the learning of high-level semantics but also supports scalable multi-task processing and modular integration.

Beyond serving as model components, latent DWMs also significantly augment training environments. By leveraging parallel rollouts, Think2Drive~\cite{li2024think2drive} and Raw2Drive~\cite{yang2025raw2drive} simulate hundreds of potential trajectories at each simulation step, yielding forward-looking, dense feedback signals. From another perspective, Popov et al.~\cite{popov2024mitigating} introduce stochastic disturbances within closed-loop driving scenarios, thereby training the agent to recover from accumulated shifts robustly.

In addition, Latent DWMs demonstrate significant potential for incorporating reasoning paradigms~\cite{guo2025deepseek}. By enabling flexible reasoning and rethinking within the latent space, these approaches~\cite{tan2025latent,liao2025think} can effectively tackle challenging tasks, such as ambiguous instructions and complex multi-agent interactions. This capability leads to safer and more forward-looking planning, while mitigating the visual information loss and computational inefficiency often associated with text-based Chain-of-Thought strategies.

Collectively, latent DWMs are creatively integrated with a diverse array of AD paradigms. Functioning variously as auxiliary training objectives, model components, or closed-loop simulation environments, they have introduced a significant shift in existing research paradigms. However, the divergence between latent prediction targets and physical sensory observations limits model interpretability. Although this abstraction favors strong correlations with specific tasks, it simultaneously introduces the risk of representation degeneration, where features may become detached from physical reality. While some approaches have acknowledged and tried to address this issue~\cite{li2024think2drive,yang2025raw2drive,liao2025think}, developing mechanisms to ground latent predictions effectively remains a critical imperative for future investigation.

\subsection{DWM in Vectorized Space}
\label{sec:vector}
The vectorized space comprises low-dimensional scene representations (e.g., traffic maps) synthesized using privileged data. In this paradigm, raw sensory processing is bypassed. Instead, traffic participants are abstracted as geometric primitives with specific dimensions, while traffic indications are encoded as discrete control signals. The inherent sparsity of this low-dimensional representation empowers DWMs to effectively model the diverse behaviors and complex multi-agent interactions within the scene.

The first pathway focuses on generative traffic simulation. Traditional rule-based traffic simulations~\cite{dosovitskiy2017carla} struggle to adaptively adjust behaviors, often exhibiting rigid and unrealistic patterns, and are prone to accumulating errors in closed-loop simulations due to their lack of self-correcting capacity. In contrast, Trafficbots~\cite{zhang2023trafficbots} enables the simulation of autonomous behaviors for multi-agent systems by endowing each agent with a private attribute context and a shared scene context. Recently, the application of diffusion models~\cite {rombach2022high,li2024drivingdiffusion,peebles2023scalable} provides a new impetus for closed-loop generative traffic simulation~\cite{jiang2024scenediffuser,lin2025causal,rowe2025scenario,sheng2025talk2traffic,tan2025scenediffuser++}, propelling the field toward large-scale, diverse, and precisely controllable developing directions.

Another pathway tends to infer multi-agent behavior from real-world observation and enhance ego planning. AdaptiveDriver~\cite{vasudevan2024planning} predicts the behavioral traits of each agent individually to derive future trajectories, thereby empowering rule-based planners. In a different approach, CarFormer~\cite{hamdan2024carformer} models holistic driving context information within a unified Autoregressive Transformer framework, integrating the features of each object into separate slots~\cite{wu2022slotformer}. In addition to modeling the behavior of individual agents, PIWM~\cite{gao2024dream} constructs a multi-agent interaction mechanism, which more accurately characterizes inter-agent interactions and their impact on the ego vehicle.

Distinct from real-world approaches that aim for scene fidelity and physical simulation, DWMs in low-dimensional vectorized space emphasize behavioral fidelity and intent simulation. The success of these frameworks demonstrates the existence of intrinsic regularities in driving behavior that can be internalized. The future integration of these models with physical-world DWMs constitutes a promising path for constructing comprehensive, closed-loop end-to-end DWMs.
\section{DWM Application}
\label{sec:4}

DWMs are highly flexible and widely used in various research areas, with their controllable generation enabling them to serve as effective neural simulators and data generators. By learning real-world dynamics, these models can adapt to different tasks or even combine generation and planning in a single framework. Furthermore, DWMs help traditional planners achieve better long-term results by overcoming the limits of immediate observations. This section provides a categorized overview of various DWM applications in the field of AD.

\begin{figure}[t]
    \centering
    \includegraphics[width=0.98\linewidth]{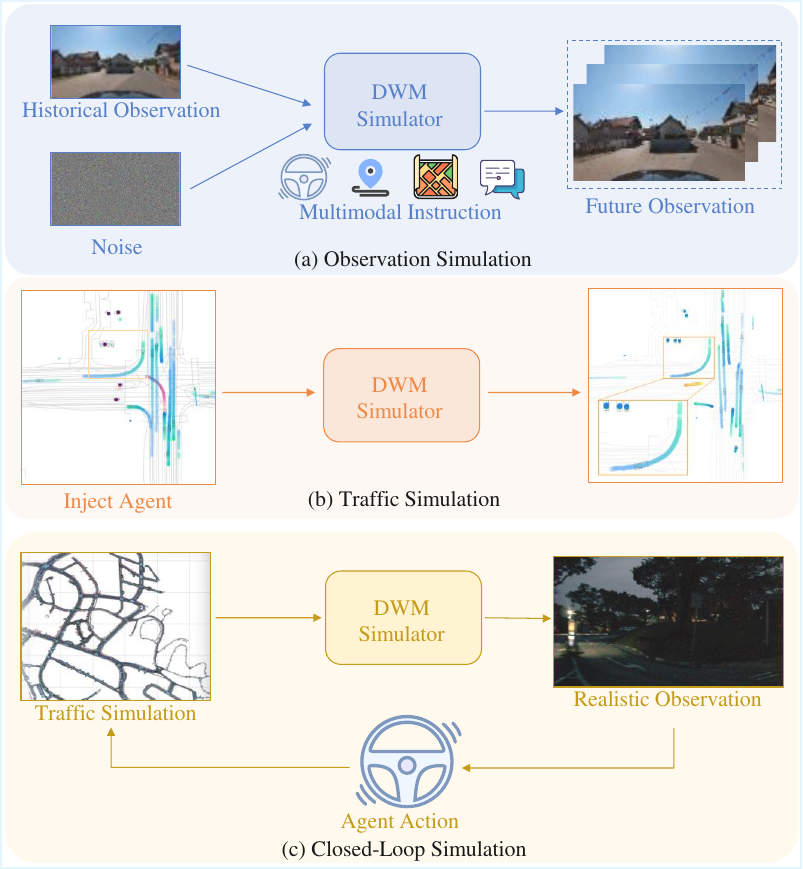}
    \caption{DWM-based simulation. (a) Future scene simulation following multimodal instruction. (b) Generative traffic simulation and editing. (c) Closed-loop simulation providing photorealistic observation. }
    \label{fig:simulation}
\end{figure}

\begin{figure*}[t]
    \centering
    \includegraphics[width=0.98\linewidth]{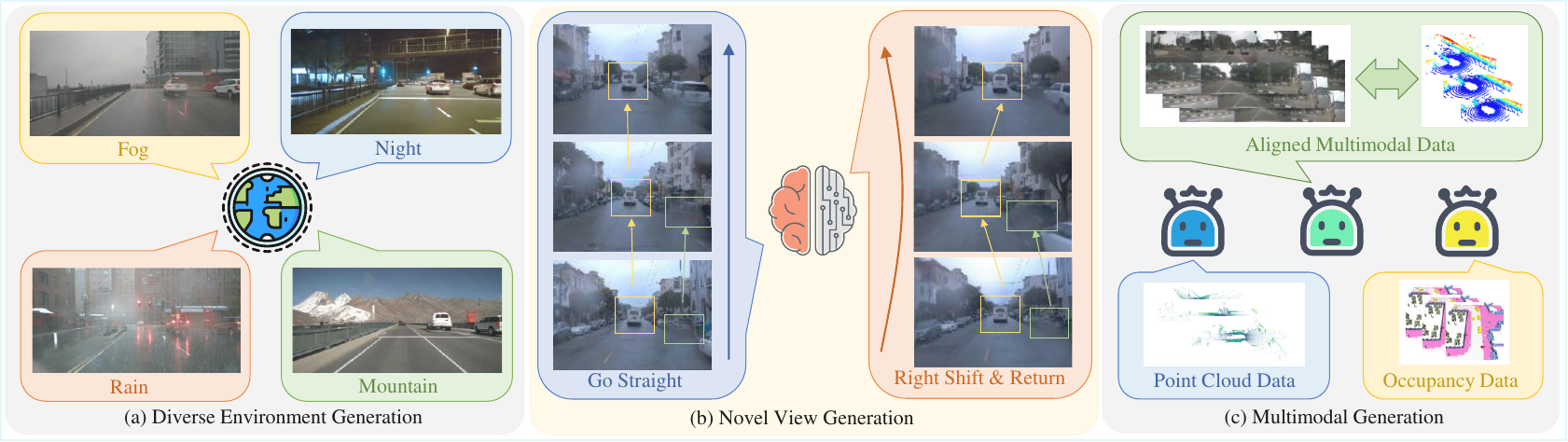}
    \caption{DWM-based data generation. (a) DWM generates data of the same layouts but diverse environments. (b) DWM generates novel views of the same scene. (c) Different DWMs support the generation of various data modalities. }
    \label{fig:generation}
\end{figure*}

\subsection{Simulation} 
Simulators constitute a pivotal instrument within the AD domain, affording a mechanism to simulate reproducible and complex driving scenarios within virtual environments. This significantly attenuates training and evaluation costs, enhances safety, and establishes a platform for equitable benchmarking. While engine-based simulators, such as CARLA~\cite{dosovitskiy2017carla} and NAVSIM~\cite{dauner2024navsim}, have substantially propelled advancements in AD research, they remain constrained by their reliance on fixed digital assets, suboptimal photorealism, and limited diversity. The potent generative capabilities of DWMs hold promise for addressing the deficiencies inherent in traditional simulators, delineating the trajectory for the next generation of generative simulators.

Extensive research across diverse modalities, including video~\cite{gao2024vista,yang2025resim,gao2023magicdrive,gao2025magicdrive}, point cloud~\cite{zyrianov2024lidardm,liang2025lidarcrafter,zhang2023learning,zhou2025lagen}, and occupancy~\cite{wang2024occsora,gu2024dome}, has demonstrated that DWMs possess robust generative capabilities and instruction-following proficiency, thereby establishing a solid foundation for high-fidelity simulation. Through training with massive real-world data, these approaches can synthesize complex yet reasonable object distributions and visual textures that remain elusive for conventional graphics-based simulators. 

As illustrated in Fig.~\ref{fig:simulation}(a), DWM-based simulators synthesize dynamic driving environments directly from raw observations~\cite{hu2023gaia,gao2024vista,zhang2023learning,zheng2024occworld,li2025driverse} and structural conditions~\cite{zyrianov2024lidardm,wang2023drivedreamer,wang2024occsora,gao2025magicdrive,gao2023magicdrive}, in contrast to conventional engine-based simulators that necessitate laborious manual asset creation, which suffer from high development costs and alleviates the constraints of limited scenario diversity. DWM-based simulators accommodate a diverse spectrum of control modalities and granularities. For instance, Vista~\cite{gao2024vista} and Drive-OccWorld~\cite{yang2024driving} accept high-level intent directives such as navigational commands, alongside low-level behavioral constraints including trajectories, steering angles, and speeds, GEM~\cite{hassan2024gem} enables precise manipulation of object distributions and pedestrian poses, while DriveDreamer-2~\cite{zhao2024drivedreamer}, Dreamland~\cite{mo2025dreamland}, and Talk2Traffic~\cite{sheng2025talk2traffic} further facilitate interactive scene editing via language prompts. In addition, DWMs support the synthesis of details unspecified by instructions, ensuring that induced transitions are both coherent and physically plausible. The cost-effectiveness, ease of use, and flexible controllability inherent in this generative paradigm significantly streamline the simulation pipeline, lowering the barrier to entry sufficiently to permit usage by non-expert personnel.

Traditional simulators typically rely on vehicle trajectories harvested from real-world driving data. This dependence not only incurs high acquisition costs and struggles to cover long-tail distributions but also introduces domain gaps when compositing trajectories with simulated backgrounds, thereby compromising the fidelity of the simulation environment. DWMs present a promising avenue for mitigating these limitations. As shown in Fig.~\ref{fig:simulation}(b), generative traffic simulation~\cite{zhang2023trafficbots,jiang2024scenediffuser,sheng2025talk2traffic,lin2025causal,rowe2025scenario} is leveraged to endow all agents within a scene with plausible driving behaviors. SceneDiffuser++~\cite{tan2025scenediffuser++} further escalates the simulation scale to the city level, encompassing the full driving duration, whereas MARL-CCE~\cite{qiao2024modelling} focuses on modeling competitive driving interactions. Furthermore, As shown in Fig.~\ref{fig:simulation}(c), integrating traffic simulators with scene simulators~\cite{yan2025drivingsphere,yang2025drivearena} facilitates the construction of closed-loop driving, thereby significantly enhancing both the degrees of freedom and the realism of the simulation.

Fundamentally, DWM-based simulators function as learnable models rather than static assemblages of fixed digital assets and predefined behaviors. This distinction significantly broadens their applicability within AD research. Possessing inherent world knowledge and predictive capabilities, these models facilitate the direct evaluation of trajectories~\cite{gao2024vista} and the estimation of training rewards~\cite{yang2025resim,li2024think2drive,yang2025raw2drive}, while also enabling the forecasting of environmental feedback over future horizons~\cite{garg2024imagine}. Furthermore, the differentiability of their outputs permits comprehensive supervision by aligning predictions with observations~\cite{wang2025adawm,li2024enhancing,popov2024mitigating}. Their support for parallel rollouts enables the efficient simulation of the stochasticity inherent in driving, thereby yielding rich and dense supervisory signals~\cite{li2024think2drive,yang2025raw2drive,popov2024mitigating}. Finally, DWMs can be jointly trained with a planner to construct customized environments, then effectively distill teacher-planner knowledge into students~\cite{zheng2025coirl}.

In summary, DWMs not only establish a simulation pipeline characterized by superior fidelity, cost-efficiency, and controllability but also expand the frontiers of simulation tasks, fostering a deeper synergy with AD research. However, the reliability issues inherent in generative simulation constitute a significant bottleneck that restricts its widespread adoption. Although recent works~\cite{yang2025resim} have attempted to address this challenge, achieving a definitive solution remains an open problem that requires further exploration.

\begin{figure*}[t]
    \centering
    \includegraphics[width=0.98\textwidth]{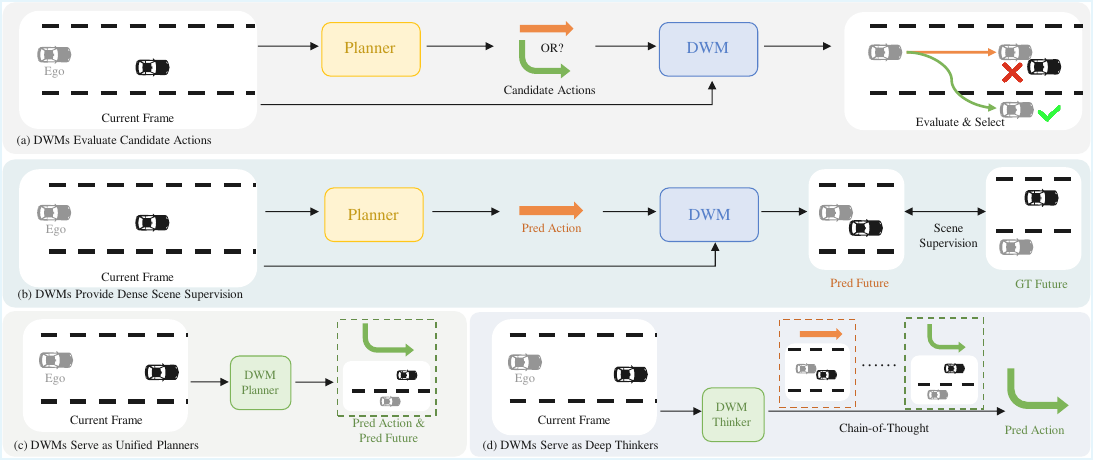}
    \caption{DWMs enhance driving. (a) DWMs evaluate candidate actions and choose the best one. (b) DWMs provide dense scene supervision for the planner through back-propagation. (c) DWMs unify prediction and planning. (d) DWMs enable reasoning with future prediction.}
    \label{fig:driving}
\end{figure*}

\begin{figure}[t]
    \centering
    \includegraphics[width=0.48\textwidth]{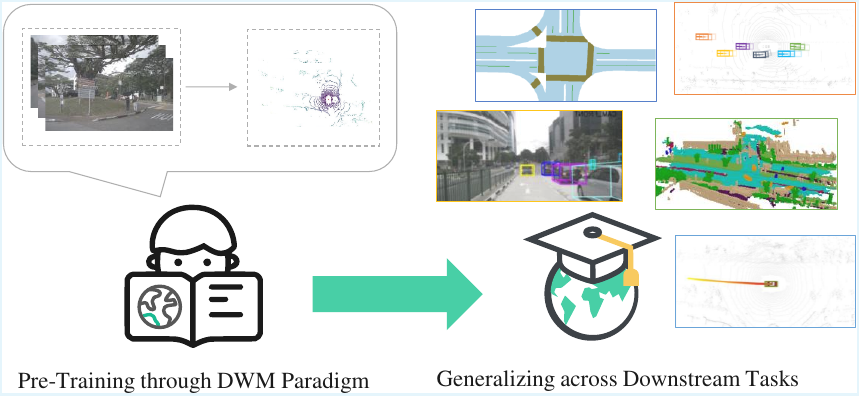}
    \caption{DWMs enable 4D pretraining. DWM learns world representation through forecasting the future, then improves performance on downstream tasks.}
    \label{fig:pretraining}
\end{figure}

\subsection{Data Generation} 
Data serves as the cornerstone of neural network research, with real-world datasets such as KITTI~\cite{Geiger2013IJRR}, Waymo~\cite{sun2020scalability}, and nuScenes~\cite{caesar2020nuscenes} standing as pivotal pillars in the AD domain. However, due to the prohibitive costs and inherent risks associated with data collection, existing datasets suffer from significant limitations. Taking the nuScenes dataset as an example, its total duration of merely 5.5 hours is significantly shorter than the empirical experience of an average human driver. Furthermore, it exhibits a highly skewed data distribution, with approximately 70\% of scenarios featuring straight lines. It also lacks sufficient diversity, failing to capture extreme scenarios such as dangerous maneuvers or traffic accidents. These deficiencies severely impede the advancement of AD research. In response, DWMs explore a cost-effective method to mitigate these challenges.

As illustrated in Fig.~\ref{fig:generation}(a), DWMs can synthesize diverse driving videos conditioned on a unified set of structured inputs~\cite{wang2023drivedreamer,zhao2024drivedreamer,yang2024physical,wen2024panacea,li2024drivingdiffusion}. This effectively mitigates the scarcity of long-tail scenarios in real-world datasets, such as nighttime, rainy weather, and diverse cityscapes. Furthermore, this synthetic data can be directly utilized to train downstream tasks, yielding significant performance improvements. A distinct paradigm shown by Fig.~\ref{fig:generation}(b) focuses on novel views. Some approaches~\cite{lu20254d,ost2025lsd,gao2024magicdrive3d} directly employ DWMs to synthesize 3D Gaussian Splatting (3D-GS)~\cite{kerbl20233d} scenes that enable arbitrary viewpoint sampling, while others~\cite{ni2024recondreamer,zhao2024drivedreamer4d} utilize DWMs to generate video data featuring diverse driving behaviors, thereby facilitating the training of 3D-GS reconstruction frameworks. In parallel, DiST-4D~\cite{guo2025dist} utilizes a decoupled spatiotemporal diffusion framework to achieve both future temporal prediction and arbitrary view synthesis. Additionally, TeraSim-World~\cite{wang2025terasim} supports the construction of diverse adversarial scenarios at arbitrary geographic locations, effectively generating corner cases and mitigating the data scarcity inherent in long-tail distributions.

The acquisition cost of 3D data, such as point clouds and occupancy grids, is significantly higher than that of driving videos. In addition to content distribution skew, driving datasets suffer from severe modal imbalance. DWMs illustrated by Fig.~\ref{fig:generation}(c) effectively lower the barrier to acquiring 3D data. Point-cloud-targeted paradigms~\cite{zyrianov2024lidardm,liang2025lidarcrafter,hu2024rangeldm} support the generation of realistic scene appearance and faithfully simulate sensor noise. Meanwhile, occupancy-targeted DWMs~\cite{wang2024occsora,yan2024renderworld,bian2024dynamiccity} enable the synthesis of scalable, dense 4D scenes enriched with semantic labels. Finally, multimodal DWMs~\cite{wu2024holodrive,liang2025seeing,li2025uniscene,alhaija2025cosmos,wu2025moviedrive} facilitate the generation of high-quality aligned multimodal data or the translation of existing modalities, thereby supporting self-supervised learning paradigms and large-scale research into multi-sensor AD.

Constrained by prohibitive costs and inherent safety risks, real-world driving datasets struggle to achieve a balanced and diverse distribution across scenarios, behaviors, and modalities. In contrast, DWMs offer a promising, cost-effective path to address this data bottleneck. They not only synthesize critical corner cases to enhance data coverage but also serve to diversify and augment existing datasets. Nevertheless, generating high-resolution scenes remains computationally intensive, and purely synthetic data cannot yet fully replace real-world data. Consequently, improving synthesis efficiency and further elevating data quality remain critical challenges to be addressed.

\subsection{Enhancing Driving}
Conventional AD paradigms typically plan actions relying solely on historical context. Lacking an understanding of the consequences of their actions and the future temporal evolution of the scene, these models suffer from a deficiency in forward-looking planning and risk assessment. Although approaches such as UniAD~\cite{hu2023planning} and VAD~\cite{jiang2023vad} incorporate prediction modules, they remain confined to predicting traffic participants. This approach not only necessitates intricate annotations of various aux-tasks but also overlooks the influence of ego-motion on the global observational environment. In contrast, the introduction of DWMs empowers planners to gain a deeper understanding of the relationship between ego-behavior and environmental dynamics, thereby enhancing the safety and foresight of driving.

The most straightforward strategy to enhance planning involves integrating a DWM directly into the planning pipeline. As shown in Figure~\ref{fig:driving}(a), the planner generates multiple candidate trajectories, for which the DWM predicts the resulting scene evolution. Rewards or costs are then derived from these predicted futures to select the optimal decision. While similar approaches~\cite{li2025end,wang2024driving,yang2024driving,li2025imagidrive} yield significant gains in planning performance, the requirement for multiple planning and prediction iterations at each step severely impedes the runtime efficiency of the AD system. Furthermore, this sample-and-filter paradigm essentially shifts the planning task from a pass@1 to an approximate pass@K regime, masking the planner's deficiencies rather than enhancing its intrinsic capabilities. In contrast, PWM~\cite{zhao2025forecasting} and SeerDrive~\cite{zhang2025future} utilize DWMs to predict future scenes prior to planning, enhancing the planner's foresight by extending the input context.
Furthermore, as illustrated in Figure~\ref{fig:driving}(b), other approaches~\cite{popov2024mitigating,li2024enhancing} integrate planning outputs into the training pipeline. By aligning predicted scenes with GT observations, these methods provide denser supervision for the planner, thereby improving its performance without imposing additional computational overhead during inference.

As presented by Fig.~\ref{fig:driving}(c), unifying planning and prediction tasks represents a more elegant solution, adhering closely to the end-to-end philosophy. Autoregressive Transformer frameworks~\cite{zheng2024occworld,hu2024drivingworld,hu2023gaia,chen2024drivinggpt,russell2025gaia} unify these tasks under the next-token prediction paradigm, directly internalizing the comprehension of dynamic driving scenes into the model itself to enhance planning capabilities. Recent advancements~\cite{wei2024occllama,zheng2024doe,zhou2025hermes,xu2025occ} extend this by establishing a unified prediction-planning-understanding framework, where textual understanding tasks further reinforce scene cognition and improve the interpretability of decision-making behaviors. Additionally, DriveLaw~\cite{xia2025drivelaw} further integrates prediction and planning within a unified representation. Finally, Fig.~\ref{fig:driving}(d) illustrates the pathway to integrate this unified architecture with Chain-of-Thought reasoning~\cite{tan2025latent,zeng2025futuresightdrive}, enhancing planning performance while enabling reflection and trial-and-error, thus demonstrating a promising trajectory for the development of DWM-based unified frameworks.

In summary, DWMs are facilitating a paradigm shift in AD from direct reactive planning to foresight-driven scene cognition. While integrating DWMs as auxiliary modules significantly enhances performance, constructing a unified model represents the definitive future trajectory for end-to-end AD. However, unified architectures remain hindered by the gaps between distinct tasks and have not yet established a significant advantage over traditional methods. Therefore, designing flexible unified architectures capable of bridging these gaps and unleashing the potential of DWMs remains a pivotal research direction.

\subsection{Pre-training}
Mainstream perception techniques, such as object detection, tracking, and depth estimation, are confined to learning representations of specific objects or isolated perspectives. Furthermore, their reliance on intricate annotations significantly impedes the effective utilization of large-scale datasets. In contrast, DWMs can capture faithful representations of the real world and learn the physical laws governing world dynamics~\cite{hu2023gaia} through learning to forecast scene evolution. Thus, pre-training on DWM tasks equips the model with comprehensive environmental cognition, enabling robust generalization performance across a diverse range of downstream tasks.

Vision-only frameworks constitute a pivotal research trajectory within the domain of embodied intelligence, encompassing AD. However, recent research~\cite{yang2025thinking,yang2025cambrian} has demonstrated that constructing a comprehensive cognition of physical space solely from visual inputs presents substantial challenges. This difficulty is particularly acute in high-speed driving scenarios, where the precise perception of 3D geometry and temporal dynamics remains exceptionally demanding while paramount for driving safety. To address this challenge, DWMs~\cite{min2023uniworld,hu2024drivingworld,yang2024visual} establish a vision-to-4D predictive pre-training paradigm illustrated by Fig.~\ref{fig:pretraining}. By leveraging self-supervised learning on large-scale image-4D data pairs, these models effectively construct 4D dynamic representations. This approach has yielded significant performance gains across multiple downstream tasks, including 3D object detection~\cite{li2024bevformer}, semantic occupancy prediction~\cite{tong2023scene}, and driving planning~\cite{hu2023planning}, validating the robust generalization capability of DWMs. Similarly, within the domain of 3D-based framework research~\cite{agro2024uno,diehl2025dio,zhu2025ad}, DWMs likewise exhibit superior transferability. Furthermore, multimodal DWMs~\cite{huang2024neural,zhang2024bevworld,shi2025drivex} jointly capture the geometric structure and visual semantic information of the environment. This capability mitigates modality bias, leading to a more holistic cognition and facilitating a broader spectrum of applications.

Although the DWM-based pre-training paradigm has demonstrated robust generalizability and high data efficiency, its further expansion is impeded by the limited scale of aligned multimodal data. To address this, PreWorld~\cite{li2025semi} explores a pathway for pre-training spatial representations using abundant 2D data. Concurrently, other DWM studies are investigating the generation of aligned multimodal data~\cite{wu2024holodrive,li2025scaling,ren2025cosmos}. Consequently, how to surmount these data constraints remains a pivotal direction for future research.
\begin{table}[t]
    \tiny
    \centering
    \setlength{\tabcolsep}{1.3mm}
    \caption{Video generation performance on the nuScenes~\cite{caesar2020nuscenes} validation set. MV means supporting multi-view forecasting. Core Arc. denotes the core architecture of the DWM. Auto. Tran. is Autoregressive Transfomer. MD means the max generation duration.}
    \begin{tabular}{lcccccccccc}
    \toprule
    Method & Reference & MV &  Resolution & Core Arc. & FID $\downarrow$ & FVD $\downarrow$ & MD (s) \\
    \midrule
    MagicDrive~\cite{gao2023magicdrive} & ICLR 24 & \ding{51} & 224×400 & Diffusion & 16.2 & - & - \\
    DriveDreamer~\cite{wang2023drivedreamer} & ECCV 24 & \ding{55} & 256×448 & Diffusion & 14.9 & 340.8 & 4 \\
    DrivingDiffusion~\cite{li2024drivingdiffusion} & ECCV 24 & \ding{51} & 512×512 & Diffusion & 15.8 & 332.0 & - \\
    Panacea~\cite{wen2024panacea} & CVPR 24 & \ding{51} & 256×512 & Diffusion & 17.0 & 139.0 & - \\
    Drive-WM~\cite{wang2024driving} & CVPR 24 & \ding{51} & 192×384 & Diffusion & 15.8 & 122.7 & 8 \\
    
    WoVoGen~\cite{lu2024wovogen} & ECCV 24 & \ding{51} & 256×448 & Diffusion & 27.6 & 417.7 & 3 \\
    GenAD~\cite{yang2024genad} & CVPR 24 & \ding{55} & - & Diffusion & 15.4 & 184.0 & 4 \\
    Vista~\cite{gao2024vista} & NeurIPS 24 & \ding{55} & 576×1024 & Diffusion & 6.9 & 89.4 & 15 \\
    BEVWorld~\cite{zhang2024bevworld} & arXiv 24 & \ding{51} & - & Diffusion & 19.0 & 154.0 & - \\
    InfinityDrive~\cite{guo2024infinitydrive} & arXiv 24 & \ding{55} & 576×1024 & Diffusion & 10.9 & 70.1 & 120 \\
    HoloDrive~\cite{wu2024holodrive} & arXiv 24 & \ding{51} & - & Diffusion & 13.6 & 103.0 & - \\
    DrivePhysica~\cite{yang2024physical} & arXiv 24 & \ding{51} & - & Diffusion & 4.0 & 38.1 & - \\
    UniMLVG~\cite{chen2024unimlvg} & arXiv 24 & \ding{51} & - & Diffusion & 5.8 & 36.1 & 20 \\
    DrivingWorld~\cite{hu2024drivingworld} & arXiv 24 & \ding{55} & 512×1024 & Auto. Tran. & 7.4 & 90.9 & 40 \\
    Doe-1~\cite{zheng2024doe} & arXiv 24 & \ding{55} & 384×672 & LLM & 15.9 & - & - \\
    DriveDreamer-2~\cite{zhao2024drivedreamer} & AAAI 25 & \ding{51} & 256×448 & Diffusion & 11.2 & 55.7 & - \\
    UniScene (GT Occ)~\cite{li2025uniscene} & CVPR 25 & \ding{51} & 256×512 & Diffusion & 6.1 & 70.5 & - \\
    GEM~\cite{hassan2024gem} & CVPR 25 & \ding{55} & 320×576 & Diffusion & 10.5 & 158.5 & 15 \\
    MaskGWM-long~\cite{ni2025maskgwm} & CVPR 25 & \ding{55} & - & Diffusion & 4.0 & 59.4 & - \\
    MagicDrive-V2~\cite{gao2025magicdrive} & ICCV 25 & \ding{51} & 848×1600 & Diffusion & 20.9 & 94.8 & 24\\
    DiST-T~\cite{guo2025dist} & ICCV 25 & \ding{51} & 424×800 & Diffusion & 6.8 & 22.7 & -\\
    Epona~\cite{zhang2025epona} & ICCV 25 & \ding{55} & 512×1024 & Diffusion & 7.5 & 82.8 & 120 \\
    DriVerse~\cite{li2025driverse} & ACM MM 25 & \ding{55} & 480×832 & Diffusion & 18.2 & 95.2 & 15 \\
    OmniGen~\cite{tang2025omnigen} & ACM MM 25 & \ding{51} & - & Diffusion & 22.2 & - & - \\  
    STAGE~\cite{wang2025stage} & IROS 25 & \ding{55} & 768×512 & Diffusion & 11.0 & 242.8 & - \\
    MiLA~\cite{wang2025mila} & arXiv 25 & \ding{51} & - & Diffusion & 3.0 & 18.2 & 10 \\
    CoGen~\cite{ji2025cogen} & arXiv 25 & \ding{51} & 360×640 & Diffusion & 10.2 & 68.4 & \\
    GeoDrive~\cite{chen2025geodrive} & arXiv 25 & \ding{55} & - & Diffusion & 4.1 & 61.6 & - \\
    MoVieDrive~\cite{wu2025moviedrive} & arXiv 25 & \ding{51} & 512×256 & Diffusion & - & 46.8 & - \\
    UniScenev2~\cite{li2025scaling} & arXiv 25 & \ding{51} & - & Diffusion & 7.6 & 61.4 & - \\
    CVD-STORM~\cite{zhang2025cvd} & arXiv 25 & \ding{51} & 256×448 & Diffusion & 3.8 & 14.0 & 20 \\
    UniFuture~\cite{liang2025seeing} & arXiv 25 & \ding{55} & 320×576 & Diffusion & 11.8 & 99.9 & - \\
    LongDWM~\cite{wang2025longdwm} & AAAI 26 & \ding{55} & 720×480 & Diffusion & 12.3 & 102.9 & 11 \\ 
    \bottomrule   
    \end{tabular}
    \label{tab:video}
\end{table}

\begin{table}[t]
    \scriptsize
    \centering
    \setlength{\tabcolsep}{0.9mm}
    \caption{Point cloud forecasting prtformance on the nuScenes~\cite{caesar2020nuscenes} validation set. Core Arc. denotes the core architecture of the DWM. Avg. corresponds to the mean of the scores at 1s, 2s, and 3s. L and C separately denote Lidar and Camera.}
    \begin{tabular}{lccccccc}
    \toprule
    \multirow{2.3}{*}{Method} & \multirow{2.3}{*}{Reference} & \multirow{2.3}{*}{Core Arc.} & \multirow{2.3}{*}{Input} &\multicolumn{4}{c}{Chamfer Distance ($m^{2}$) $\downarrow$} \\
    \cmidrule(lr){5-8}
     & & & & 1s & 2s & 3s & Avg. \\
    \midrule
    Copilot4D~\cite{zhang2023learning} & ICLR 24 & Diffusion & L & 0.36 & - & 0.58 & - \\
    LiSTAR~\cite{liu2025listar} & arXiv 25 & Transformer & L & 0.30 & - & - & - \\
    LaGen~\cite{zhou2025lagen} & arXiv 25 & Diffusion & L & 0.61 & 0.81 & 1.08 & 0.83 \\
    \midrule
    ViDAR~\cite{yang2024visual} & CVPR 24 & Transformer & C & 1.12 & 1.38 & 1.73 & 1.41 \\
    HERMES~\cite{zhou2025hermes} & ICCV 25 & LLM & C & 0.78 & 0.95 & 1.17 & 0.97 \\
    DriveX-B~\cite{shi2025drivex} & ICCV 25 & Transformerr & C & 0.66 & 0.86 & 1.10 & 0.87 \\
    \midrule
    BEVWorld~\cite{zhang2024bevworld} & arXiv 24 & Diffusion & L \& C & 0.44 & - & 0.73 & - \\
    HoloDrive~\cite{wu2024holodrive} & arXiv 24 & Diffusion & L \& C & - & - & 0.83 & - \\
    \bottomrule   
    \end{tabular}
    \label{tab:point_cloud}
\end{table}

\begin{table*}[ht]
    \scriptsize
    \centering
    \setlength{\tabcolsep}{2.1mm}
    \caption{Occupancy forecasting performance on Occ3D-nuScenes~\cite{tian2023occ3d}. Core Arc. denotes the core architecture of the DWM. Auto. Tran. is Autoregressive Transfomer. Recon. denotes the score at current time step. Avg. corresponds to the mean of the scores at 1s, 2s, and 3s. O and C separately denote Occupancy and Camera.}
    \begin{tabular}{lccccccccccccc}
    \toprule 
    \multirow{2.3}{*}{Method} & \multirow{2.3}{*}{Reference} & \multirow{2.3}{*}{Core Arc.} & \multirow{2.3}{*}{Input} & \multicolumn{5}{c}{mIoU (\%) $\uparrow$} & \multicolumn{5}{c}{IoU  (\%) $\uparrow$} \\
     \cmidrule(lr){5-9}
     \cmidrule(lr){10-14}
     & & & & Recon. & 1s & 2s & 3s & Avg. & Recon. & 1s & 2s & 3s & Avg. \\
      \midrule
    OccWorld-O~\cite{zheng2024occworld} & ECCV 24 & Auto. Tran. & O & 66.38 & 25.78 & 15.14 & 10.51 & 17.14 & 62.29 & 34.63 & 25.07 & 20.18 & 26.63 \\
    OccSora~\cite{wang2024occsora} & arXiv 24 & Diffusion & O & 27.40 & - & - & - & - & 37.00 & - & - & - & - \\
    OccLLaMA-O~\cite{wei2024occllama} & arXiv 24 & LLM & O & 75.20 & 25.05 & 19.49 & 15.26 & 19.93 & 63.76 & 34.56 & 28.53 & 24.41 & 29.17 \\
    DFIT-OccWorld-O~\cite{zhang2024efficient} & arXiv 24 & Auto. Tran. & O & - & 31.68 & 21.29 & 15.18 & 22.71 & - & 40.28 & 31.24 & 25.29 & 32.27 \\
    DOME-O~\cite{gu2024dome} & arXiv 24 & Diffusion & O & 83.08 & 35.11 & 25.89 & 20.29 & 27.10 & 77.25 & 43.99 & 35.36 & 29.74 & 36.36 \\
    RenderWorld~\cite{yan2024renderworld} & ICRA 25 & Auto. Tran. & O & - & 28.69 & 18.89 & 14.83 & 20.80 & - & 37.74 & 28.41 & 24.08 & 30.08 \\
    Occ-LLM~\cite{xu2025occ} & ICRA 25 & LLM & O & - & 36.65 & 32.14 & 28.77 & 32.52 & - & 24.02 & 21.65 & 17.29 & 20.99 \\
    DynamicCity~\cite{bian2024dynamiccity} & ICLR 25 & Diffusion & O & 40.79 & - & - & - & - & - & - & - & - & - \\
    $I^{2}$-World~\cite{liao2025i2} & ICCV 25 & Transformer & O & 81.22 & 47.62 & 38.58 & 32.98 & 39.73 & 68.30 & 54.29 & 49.43 & 45.69 & 49.80 \\
    Liu. et. al.~\cite{liu2025towards} & NeurIPS 25 & Transformer & O & - & 36.42 & 27.39 & 21.66 & 28.49 & - & 43.68 & 36.89 & 31.98 & 37.52 \\
    COME~\cite{shi2025come} & arXiv 25 & Diffusion & O & - & 42.75 & 32.97 & 26.98 & 34.23 & - & 50.57 & 43.47 & 38.36 & 44.13 \\
    OccTENS-O~\cite{jin2025occtens} & arXiv 25 & Auto. Tran. & O & - & 27.96 & 21.75 & 16.47 & 22.06 & - & 38.73 & 29.50 & 24.86 & 31.03 \\
    DTT-O~\cite{xu2025temporal} & arXiv 25 & Transformer & O & 85.50 & 37.69 & 29.77 & 25.10 & 30.85 & 92.07 & 76.60 & 74.44 & 72.71 & 74.58 \\
    \midrule
    OccWorld-S~\cite{zheng2024occworld} & ECCV 24 & Auto. Tran. & C & 0.27 & 0.28 & 0.26 & 0.24 & 0.26 & 4.32 & 5.05 & 5.01 & 4.95 & 5.00 \\
    OccWorld-D~\cite{zheng2024occworld} & ECCV 24 & Auto. Tran. & C & 18.63 & 11.55 & 8.10 & 6.22 & 8.62 & 22.88 & 18.90 & 16.26 & 14.43 & 16.53 \\
    OccLLaMA-F~\cite{wei2024occllama} & arXiv 24 & LLM & C & 37.38 & 10.34 & 8.66 & 6.98 & 8.66 & 38.92 & 25.81 & 23.19 & 19.97 & 22.99 \\
    DOME-F~\cite{gu2024dome} & arXiv 24 & Diffusion & C & 75.00 & 24.12 & 17.41 & 13.24 & 18.25 & 74.31 & 35.18 & 27.90 & 23.44 & 28.84 \\
    DFIT-OccWOrld-V~\cite{zhang2024efficient} & arXiv 24 & Auto. Tran. & C & - & 13.38 & 10.16 & 7.96 & 10.50 & - & 19.18 & 16.85 & 15.02 & 17.02 \\
    PreWorld~\cite{li2025semi} & ICLR 25 & Transformer & C & - & 12.27 & 9.24 & 7.15 & 9.55 & - & 23.62 & 21.62 & 19.63 & 21.62 \\
    RenderWorld~\cite{yan2024renderworld} & ICRA 25 & Auto. Tran. & C & - & 2.83 & 2.55 & 2.37 & 2.58 & - & 14.61 & 13.61 & 12.98 & 13.73 \\
    Occ-LLM~\cite{xu2025occ} & ICRA 25 & LLM & C & - & 27.11 & 24.07 & 20.19 & 23.79 & - & 11.28 & 10.21 & 9.13 & 10.21 \\
    $I^{2}$-World~\cite{liao2025i2} & ICCV 25 & Transformer & C & 25.13 & 21.67 & 18.78 & 16.47 & 18.97 & 32.66 & 30.55 & 28.76 & 26.99 & 28.77 \\
    DTT-F~\cite{xu2025temporal} & arXiv 25 & Transformer & C & 43.52 & 24.87 & 18.30 & 15.63 & 19.60 & 54.31 & 38.98 & 37.45 & 31.89 & 36.11 \\
    COME~\cite{shi2025come} & arXiv 25 & Diffusion & C & - & 26.56 & 21.73 & 18.49 & 22.26 & - & 48.08 & 43.84 & 40.28 & 44.07 \\
    SparseWorld-TC~\cite{du2025sparseworld} & arXiv 25 & VGGT~\cite{wang2025vggt} & C & - & 32.76 & 29.62 & 27.28 & 29.89 & - & 55.28 & 53.56 & 51.71 & 53.52 \\
    OccTENS--F~\cite{jin2025occtens} & arXiv 25 & Auto. Tran. & C & - & 17.17 & 10.38 & 7.82 & 11.79 & - & 27.60 & 25.14 & 20.33 & 24.35 \\
    SparseWorld~\cite{dang2025sparseworld} & AAAI 26 & Transformer & C & - & 14.93 & 13.15 & 11.51 & 13.20 & - & 22.96 & 22.10 & 21.05 & 22.03 \\
    \midrule
    UniScene~\cite{li2025uniscene} & CVPR 25 & Diffusion & Layout & 92.10 & 35.37 & 29.59 & 25.08 & 31.76 & 87.00 & 38.34 & 32.70 & 29.09 & 34.84 \\    
    \bottomrule   
    \end{tabular}
    \label{tab:occupancy}
\end{table*}

\section{DWM Performance}
\label{sec:5}
DWMs are widely adopted in the field of AD and have significantly advanced relevant research. To better illustrate this development, this section reviews the performance of representative works across major tasks. These tasks include video, point cloud, and occupancy generation, as well as open-loop and closed-loop planning. 

\subsection{Video Prediction}
Driving video generation aims to predict future frame sequences from single or multiple views based on past observations and control signals. Generating high-quality and high-resolution results requires a balance between semantic logic and visual details. Additionally, the need for consistency across views and long-term prediction creates significant challenges for spatiotemporal dynamic modeling. Tab.~\ref{tab:video} illustrates the performance of DWMs for the driving video generation task on the nuScenes~\cite{caesar2020nuscenes} validation set.

The final generative performance is influenced by various factors. For instance, structured conditions incorporate spatial priors while command conditions do not. Meanwhile, generating high-resolution and long-term content is significantly more challenging, resulting in lower FID and FVD scores compared to low-resolution and short-term content. The comparison presented in Tab.~\ref{tab:video} is therefore not strictly fair. However, from an overall perspective, DWMs have achieved significant performance progress, delivering higher FID and FVD scores, greater resolution, and more stable long-term generation, demonstrating a major advance in the overall quality of generated video content.

It is important to note that earlier models frequently exhibited FVD values significantly higher than their FID scores~\cite{li2024drivingdiffusion,wang2023drivedreamer,lu2024wovogen}. This difference suggested a lack of ability in temporal modeling. In contrast, recent studies simultaneously reduce both FID and FVD, significantly narrowing the gap between these two metrics~\cite{wang2025longdwm,zhang2025cvd,chen2024unimlvg}. This change shows that these models have evolved from understanding single frames to achieving coherent video perception. Furthermore, improvements in resolution and generation horizon suggest that these models are steadily moving toward simulating complete driving events.

\subsection{Point Cloud Prediction}
The point cloud prediction task requires both the effective modeling of spatiotemporal dynamics and the integration of high-fidelity rendering techniques, which accurately emulate the distribution of real-world point clouds. Tab.~\ref{tab:point_cloud} presents a performance comparison of representative methods on the nuScenes~\cite{caesar2020nuscenes} validation set. It shows the Chamfer Distance values at three future time steps, where Avg. corresponds to the arithmetic mean of these three values.

The prediction performance of point-cloud-targeted DWMs depends heavily on specific experimental settings. For instance, frameworks that leverage point cloud inputs, such as Copilot4D~\cite{zhang2023learning} and LaGen~\cite{zhou2025lagen}, demonstrate superior performance compared to vision-only counterparts like ViDAR~\cite{yang2024visual} and DriveX~\cite{shi2025drivex}. Nevertheless, vision-only approaches~\cite{yang2024visual,zhou2025hermes,shi2025drivex,zhou2025lagen} are showing a steady increase in their prediction ability.

\begin{table*}[ht]
    \scriptsize
    \centering
    \setlength{\tabcolsep}{3.3mm}
    \caption{Open-loop driving performance on the nuScenes~\cite{caesar2020nuscenes} validation set. Core Arc. denotes the core architecture of the DWM. Auto. Tran. is Autoregressive Transfomer. Pred. Mod. means the modality of predicted scenes. Avg. corresponds to the mean of the scores at 1s, 2s, and 3s. O, C, and D separately denote Occupancy, Camera, and Depth.}
    \begin{tabular}{lccccccccccc}
    \toprule
    \multirow{2.3}{*}{Method} & \multirow{2.3}{*}{Reference} & \multirow{2.3}{*}{Core Arc.} & \multirow{2.3}{*}{Pred. Mod.} & \multicolumn{4}{c}{L2 ($m$) $\downarrow$} & \multicolumn{4}{c}{Col. (\%) $\downarrow$} \\
     \cmidrule(lr){5-8}
     \cmidrule(lr){9-12}
     & & & & 1s & 2s & 3s & Avg. & 1s & 2s & 3s & Avg. \\
    \midrule
    \textcolor{gray}{UniAD~\cite{hu2023planning}} & \textcolor{gray}{CVPR 23} & \textcolor{gray}{Transformer} & \textcolor{gray}{Not DWM} & \textcolor{gray}{0.48} & \textcolor{gray}{0.96} & \textcolor{gray}{1.65} & \textcolor{gray}{1.03} & \textcolor{gray}{0.05} & \textcolor{gray}{0.17} & \textcolor{gray}{0.71} & \textcolor{gray}{0.31} \\
    \textcolor{gray}{VAD~\cite{jiang2023vad}} & \textcolor{gray}{ICCV 23} & \textcolor{gray}{Transformer} & \textcolor{gray}{Not DWM} & \textcolor{gray}{0.54} & \textcolor{gray}{1.15} & \textcolor{gray}{1.98} & \textcolor{gray}{1.22} & \textcolor{gray}{0.10} & \textcolor{gray}{0.24} & \textcolor{gray}{0.96} & \textcolor{gray}{0.43} \\
    \midrule
    Drive-WM~\cite{wang2024driving} & CVPR 24 & Diffusion & C & 0.43 & 0.77 & 1.20 & 0.80 & 0.10 & 0.21 & 0.48 & 0.26 \\
    DriveDreamer~\cite{wang2023drivedreamer} & ECCV 24 & Diffusion & C & - & - & - & 0.29 & - & - & - & 0.15 \\
    Doe-1~\cite{zheng2024doe} & arXiv 24 & LLM & C & 0.50 & 1.18 & 2.11 & 1.26 & 0.04 & 0.37 & 1.19 & 0.53 \\
    Epona~\cite{zhang2025epona} & ICCV25 & Diffusion & C & 0.61 & 1.17 & 1.98 & 1.25 & 0.01 & 0.22 & 0.85 & 0.36 \\
    PWM~\cite{zhao2025forecasting} & NeurIPS 25 & LLM & C & 0.41 & 0.75 & 1.17 & 0.78 & 0.01 & 0.01 & 0.18 & 0.07 \\
    FSDrive~\cite{zeng2025futuresightdrive} & NeurIPS 25 & LLM & C & 0.18 & 0.39 & 0.77 & 0.45 & 0.00 & 0.06 & 0.42 & 0.16 \\
    \midrule
    Occ-World-O~\cite{zheng2024occworld} & ECCV 24 & Auto. Tran. & O & 0.43 & 1.08 & 1.99 & 1.17 & 0.07 & 0.38 & 1.35 & 0.60 \\
    Occ-LLaMA-O~\cite{wei2024occllama} & arXiv 24 & \multicolumn{1}{c}{LLM} & O & 0.37 & 1.02 & 2.03 & 1.14 & 0.04 & 0.24 & 1.20 & 0.49 \\
    DFIT-OccWorld-O~\cite{zhang2024efficient} & arXiv 24 & Auto. Tran. & O & 0.38 & 0.96 & 1.73 & 1.02 & 0.07 & 0.39 & 0.90 & 0.45 \\
    Drive-OccWorld~\cite{yang2024driving} & AAAI 25 & Transformer & O & 0.25 & 0.44 & 0.72 & 0.47 & 0.03 & 0.08 & 0.22 & 0.11 \\
    PreWorld~\cite{li2025semi} & ICLR 25 & Transformer & O & 0.22 & 0.30 & 0.40 & 0.31 & 0.21 & 0.66 & 0.71 & 0.53 \\
    RenderWorld~\cite{yan2024renderworld} & ICRA 25 & Auto. Tran. & O & 0.35 & 0.91 & 1.84 & 1.03 & 0.05 & 0.40 & 1.39 & 0.61 \\
    Occ-LLM~\cite{xu2025occ} & ICRA 25 & LLM & O & 0.12 & 0.24 & 0.49 & 0.28 & - & - & - & - \\
    IR-WM~\cite{mei2025vision} & arXiv 25 & Transformer & O & 0.23 & 0.51 & 0.85 & 0.53 & 0.14 & 0.20 & 0.16 & 0.17 \\
    DTT~\cite{xu2025temporal} & arXiv 25 & Transformer & O & 0.32 & 0.91 & 1.76 & 1.00 & 0.08 & 0.32 & 0.51 & 0.30 \\
    OccTENS-O~\cite{jin2025occtens} & arXiv 25 & Auto. Tran. & O & 0.39 & 1.02 & 1.96 & 1.12 & 0.08 & 0.25 & 1.12 & 0.48 \\
    SparseWorld~\cite{dang2025sparseworld} & AAAI 26 & Transformer & O & 0.19 & 0.25 & 0.36 & 0.27 & 0.11 & 0.29 & 0.46 & 0.29 \\
    \midrule
    NeMo~\cite{huang2024neural} & ECCV 24 & Transformer & O \& C & 0.35 & 0.94 & 1.77 & 1.02 & 0.11 & 0.30 & 1.37 & 0.59 \\
    DiST-4D~\cite{guo2025dist} & ICCV 25 & Diffusion & D \& C & 0.56 & 1.11 & 1.91 & 1.19 & - & - & - & - \\
    \midrule
    LAW~\cite{li2024enhancing} & ICLR 24 & Transformer & - & 0.24 & 0.46 & 0.76 & 0.49 & 0.08 & 0.10 & 0.39 & 0.19 \\
    World4Drive~\cite{zheng2025world4drive} & ICCV 25 & Transformer & - & 0.23 & 0.47 & 0.81 & 0.50 & 0.02 & 0.12 & 0.33 & 0.16 \\
    \bottomrule   
    \end{tabular}
    \label{tab:open_loop}
\end{table*}

\begin{table*}[ht]
    \scriptsize
    \centering
    \setlength{\tabcolsep}{2.3mm}
    \caption{Closed-loop evaluation on CARLA~\cite{dosovitskiy2017carla}. Core Arc. denotes the core architecture of the DWM. Pred. Mod. means the modality of predicted scenes. Effi. is the Efficiency score, while Comf. is the Comfortness score. O, C, and B separately denote Occupancy, Camera, and BEV Map.}
    \begin{tabular}{lccccccccccc}
    \toprule
    Method & Reference & RL & Core Arc. & Pred. Mod. & IS $\uparrow$ & RC (\%) $\uparrow$ & DS $\uparrow$ & SR (\%) $\uparrow$ & Effi. $\uparrow$ & Comf. $\uparrow$ & Benchmark \\
    \midrule
    \textcolor{gray}{VAD-Base~\cite{jiang2023vad}} & \textcolor{gray}{ICCV 23} & \textcolor{gray}{\ding{55}} & \textcolor{gray}{Transformer} & \textcolor{gray}{Not DWM} & \textcolor{gray}{0.74} & \textcolor{gray}{87.26} & \textcolor{gray}{64.29} & \textcolor{gray}{-} & \textcolor{gray}{-} & \textcolor{gray}{-} & \textcolor{gray}{Town05 Short} \\
    Nemo~\cite{huang2024neural} & ECCV 24 & \ding{55} & Transformer & C \& O & - & 90.12 & 75.87 & - & - & - & Town05 Short \\
    \midrule
    \textcolor{gray}{VAD-Base~\cite{jiang2023vad}} & \textcolor{gray}{ICCV 23} & \textcolor{gray}{\ding{55}} & \textcolor{gray}{Transformer} & \textcolor{gray}{Not DWM} & \textcolor{gray}{0.40} & \textcolor{gray}{75.20} & \textcolor{gray}{30.31} & \textcolor{gray}{-} & \textcolor{gray}{-} & \textcolor{gray}{-} & \textcolor{gray}{Town05 Long} \\
    MILE~\cite{bronstein2022hierarchical} & NeurIPS 22 & \ding{55} & RNN & C \& B & 0.63 & 97.40 & 61.10 & - & - & - & Town05 Long \\
    Nemo~\cite{huang2024neural} & ECCV 24 & \ding{55} & Transformer & C \& O & - & 80.98 & 42.57 & - & - & - & Town05 Long \\
    Popov. et. al.~\cite{popov2024mitigating}& ArXiv 24 & \ding{51} & GRU & - & 0.80 & 100.00 & 70.00 & - & - & - & Town05 Long \\
    LAW~\cite{li2024enhancing} & ICLR 25 & \ding{55} & Transformer & - & 0.72 & 97.80 & 70.10 & - & - & - & Town05 Long \\
    \midrule
    Think2Drive~\cite{li2024think2drive} & ECCV 24 & \ding{51} & RSSM & - & 0.73 & 78.20 & 43.80 & - & - & - & Town13 \\
    Raw2Drive~\cite{yang2025raw2drive} & NeurIPS 25 & \ding{51} & RSSM & - & 0.43 & 9.32 & 4.12 & - & - & - & Town13 \\
    \midrule
    \textcolor{gray}{ORION~\cite{fu2025orion}} & \textcolor{gray}{ICCV 25} & \textcolor{gray}{\ding{55}} & \textcolor{gray}{LLM} & \textcolor{gray}{Not DWM} & \textcolor{gray}{-} & \textcolor{gray}{-} & \textcolor{gray}{77.74} & \textcolor{gray}{54.62} & \textcolor{gray}{151.48} & \textcolor{gray}{17.38} & \textcolor{gray}{Bench2Drive~\cite{jia2024bench2drive}} \\
    Think2Drive~\cite{li2024think2drive} & ECCV 24 & \ding{51} & RSSM & - & - & - & 91.85 & 85.41 & 269.14 & 25.97 & Bench2Drive~\cite{jia2024bench2drive} \\
    WoTE~\cite{li2025end} & ICCV 25 & \ding{55} & Transformer & B & - & - & 61.71 & 31.36 & - & - & Bench2Drive~\cite{jia2024bench2drive} \\
    Raw2Drive~\cite{yang2025raw2drive} & NeurIPS 25 & \ding{51} & RSSM & - & - & - & 71.36 & 50.24 & 214.17 & 22.42 & Bench2Drive~\cite{jia2024bench2drive} \\
    SeerDrive~\cite{zhang2025future} & NeurIPS 25 & \ding{55} & Transformer & B & - & - & 58.32 & 30.17 & - & - & Bench2Drive~\cite{jia2024bench2drive} \\
    \bottomrule   
    \end{tabular}
    \label{tab:closed_loop}
\end{table*}

\subsection{Occupancy Prediction}
Occupancy forecasting requires modeling precise spatial structures and capturing complex spatiotemporal dynamics. Tab.~\ref{tab:occupancy} summarizes the performance of existing methods on the Occ3D-nuScenes~\cite{tian2023occ3d} validation set. Within this table, Recon. evaluates the occupancy reconstruction fidelity at the current timestamp, and Avg. represents the mean predictive performance across the three future time steps.

As shown in Tab.~\ref{tab:occupancy}, the evolution of occupancy-targeted DWM from early frameworks~\cite{zheng2024occworld,wei2024occllama,wang2024occsora} to recent architectures~\cite{du2025sparseworld,dang2025sparseworld,liao2025i2} reveals significant improvement in predictive performance. Early methodologies, exemplified by DOME~\cite{gu2024dome} and OccLLaMA~\cite{wei2024occllama}, exhibit a substantial performance gap where reconstruction IoU often exceeds prediction IoU by more than twofold. Furthermore, these methods suffer from a pronounced temporal decay as the prediction horizon extends from one to three seconds. In contrast, recent advancements~\cite{liao2025i2,du2025sparseworld} not only bolster both reconstruction and prediction benchmarks but also narrow the discrepancy between them. This trend underscores a pivotal breakthrough in capturing complex spatiotemporal dynamics. 

Notably, for the majority of the presented frameworks, the IoU score in predictive tasks typically surpasses the mIoU score, despite the fact that mIoU is often higher in reconstruction tasks. This discrepancy suggests that these models exhibit a lower proficiency in capturing the dynamics of small-scale objects compared to large-scale entities and background. In contrast, Occ-LLM~\cite{xu2025occ} allocates the preponderance of its modeling capacity toward dynamic object prediction and achieves an mIoU that exceeds the corresponding IoU. Finally, the performance of vision-to-occupancy prediction remains lower than results with occupancy or layout input, highlighting the formidable challenge of constructing 4D spatial cognition from visual inputs.

\subsection{Planning}
Planning tasks are primarily categorized into open-loop and closed-loop paradigms. In open-loop evaluations, the model plans future behaviors over a short horizon based on recorded observations. In contrast, closed-loop evaluations involve a dynamic process where the simulation environment provides real-time feedback, requiring the agent to engage in continuous interaction to complete the entire driving mission. While open-loop protocols leverage real-world datasets to provide photorealistic observations, closed-loop settings align more closely with the real-world driving process, thereby offering a more rigorous assessment of the agent's comprehensive driving capability.

\textbf{Open-Loop Planning.} Tab.~\ref{tab:open_loop} presents the open-loop planning performance of DWMs on the nuScenes validation set. In the table, Core Arc. denotes the core architecture of the DWM framework, while Pred. Mod. indicates the target prediction modality. The results cover a future 3-second planning horizon, with Avg. representing the average performance across the three evaluated time steps.

Open-loop planning is influenced by various factors, including input resolution and foundation models. However, from a comprehensive viewpoint, DWMs have generally surpassed traditional mainstream end-to-end frameworks~\cite{hu2023planning,jiang2023vad}, demonstrating the significant benefits of these models for AD research. It is noteworthy that Occ-LLM~\cite{xu2025occ} and SparseWorld~\cite{dang2025sparseworld} improve trajectory reasonableness by focusing on moving objects through decoupling methods. Meanwhile, PWM~\cite{zhao2025forecasting} and Drive-OccWorld~\cite{yang2024driving} utilize DWMs to enhance the foresight of planners regarding hazards, effectively reducing collision rates and demonstrating safer driving.

\textbf{Closed-Loop Planning.} As shown in Tab.~\ref{tab:closed_loop}, DWMs outperform traditional end-to-end methods in closed-loop planning within the widely used Town05 and Town13 benchmarks in CARLA~\cite{dosovitskiy2017carla}. Furthermore, Tab.~\ref{tab:closed_loop} also presents the performance on Bench2Drive~\cite{jia2024bench2drive}, which evaluates closed-loop performance across diverse driving environments. Think2Drive~\cite{li2024think2drive} achieves the best results by using privileged information from the simulator, but other DWMs still show a gap compared to mainstream methods~\cite{fu2025orion}. This suggests that there is still considerable room for improvement in closed-loop driving for DWMs. Notably, although Raw2Drive~\cite{yang2025raw2drive} achieves lower scores in DS score and SR score, its higher scores in Efficiency and Comfortness suggest that it has a more substantial capacity for smooth driving.
\section{Limitation and Future Direction}
\label{sec:6}
While research on DWMs has made significant progress, several limitations remain that may limit their full potential. Additionally, adjusting DWMs for various AD tasks is a continuous challenge. In this section, we provide a detailed discussion of current limitations and describe possible directions for future research and development.

\textbf{Data Scarcity.}
The acquisition of AD data involves prohibitive costs, particularly for long-tail and safety-critical cases. This results in the limited scale and uneven distribution of existing datasets. Although recent works~\cite{wang2024drivingdojo,yang2024generalized} have established large-scale and diverse driving video datasets, reliably acquiring synchronized multi-modal data remains a significant challenge.

While research on DWM-based data synthesis~\cite{yang2024physical,ni2024recondreamer} has made substantial progress and demonstrated effectiveness in various downstream tasks, the quality of such synthesized data is not guaranteed. We argue that future efforts should not only focus on enhancing the diversity and quality of data synthesis, but also on establishing a GT-free evaluation paradigm that encompasses driving behaviors, object distributions, and geometry-texture details. Leveraging powerful generative models and robust evaluation suites, a full-stack automated synthesis, filtering, and refining pipeline can be constructed to facilitate the creation of high-quality, large-scale multimodal driving datasets.

\textbf{Reliable Simulation.}
Although DWMs have been demonstrated to apply to closed-loop simulation~\cite{yang2025drivearena}, their reliability remains a major concern. A key problem is to ensure that DWM performance would not drop significantly when confronted with complex simulation scenarios, such as long-horizon rollouts and drastic viewpoint shifts, as well as variable driving conditions, including diverse traffic flows and varying weather conditions. This presents significant challenges to the robustness and generalization of DWMs. On the other hand, generative models suffer from intrinsic flaws that are prone to hallucinations and physically inconsistent outputs (e.g., sudden vehicle appearances, speed estimation errors). Even under nominal conditions, these issues can lead to hazardous decision outcomes. Finally, evaluations of DWM's predictive capabilities are predominantly based on offline datasets, which fail to reflect the dynamic interplay between the planner and the DWM-based simulator within more realistic closed-loop driving behaviors. Thus, investigating how to evaluate the capabilities for DWMs to serve as a simulator, as well as how to achieve reliable simulation, remains a valuable research direction.

\textbf{Task Unification.}
AD scene prediction is a multi-level modeling challenge encompassing environmental layout, motion dynamics, and driving intentions. However, current research faces a major division. DWMs that focus on detailed scene prediction often implicitly reflect behavioral intentions by predicting pixel variations, yet lack the explicit modeling of driving logic. Meanwhile, DWMs that focus on behavior simulation prioritize strategic planning but frequently overlook the fine-grained perception of complex, real-world visual environments. This disconnect hinders models from forming a unified cognition of driving scenes. Recently, the exploration of language capabilities has emerged as a key development trend in DWMs~\cite{zhou2025hermes,zheng2024doe,xu2025occ,wei2024occllama}, AD research~\cite{zhao2025extending,fu2025minddrive,nie2024reason2drive}, and even in the embodied intelligence domain~\cite{liang2026cook,zhang2025embodied,li2024cogact}. Looking ahead, building an integrated end-to-end framework that unifies prediction, planning, and understanding will be pivotal to unlocking the full potential of DWMs.

\textbf{Multi-Sensor Modeling.} 
AD systems universally rely on multi-sensor fusion schemes. Data from different modalities exhibits inherent modality-specific biases while simultaneously corresponding to the same physical world. Despite significant progress in unified multimodal representation~\cite{zhang2024bevworld, shi2025drivex}, the alignment of different modalities has received little attention.
For instance, most multi-modal generation approaches lack interaction between the generated outputs. Aligning these results could not only facilitate mutual error correction but also foster the learning of modality-agnostic world knowledge. Furthermore, acquiring matched multi-sensor data is considerably expensive. Leveraging easily accessible unaligned data (i.e., multi-modal data not corresponding to the same scene) for training could effectively expand the scale of available data.

\textbf{Efficiency.} 
Generative tasks pose significant challenges to the deployment efficiency of DWMs. The inherent complexity of 4D data leads to substantial computational overhead. Moreover, model scaling and the integration of technologies like LLMs and Chain-of-Thought exacerbate computational burdens and inference latency. While DWMs have achieved notable progress in predicting and planning tasks, balancing performance with efficiency remains a core obstacle to their real-world deployment. To address this, the latent DWM paradigm~\cite{gao2024enhance,li2024think2drive} circumvents efficiency bottlenecks by omitting scene decoding, while recent studies effectively enhance generation efficiency through scene decoupling~\cite{zuo2024gaussianworld,zhang2024efficient,mei2025vision,xu2025temporal} or sparse representation~\cite{deng2025gaussiandwm,du2025sparseworld,dang2025sparseworld}. Beyond model and representation lightweighting, reducing scene decoding frequency and implementing multi-task parallelism are also promising solutions. Such efficiency-centric innovations will continue to be a pivotal focus in future research.

\textbf{Attack and Defense.}
Adversarial attacks~\cite{wang2025black} pose a severe threat to AD safety, as they can easily trigger catastrophic traffic accidents. These attacks are typically implemented via carefully crafted, human-imperceptible adversarial perturbation patches. Since such samples are virtually non-existent in training data, detection and defense become incredibly challenging. Despite the importance of this issue, dedicated research on adversarial attacks against DWMs remains scarce. Therefore, investigating such attacks and developing effective defense strategies is of vital practical significance. Meanwhile, leveraging the predictive capability and generalization of DWMs to locate and mitigate the impact of adversarial patches represents a highly valuable research direction. Collectively, these efforts play a crucial role in ensuring the safe and reliable deployment of DWMs in the real world.

\section{Conclusion}
DWMs have been widely recognized as fundamental components in the architecture of AD systems. Their primary goal is to enhance a comprehensive understanding of dynamic environments by predicting the evolution of future scenes. This paper first introduces the datasets and evaluation metrics commonly used in DWM research. We then provide a systematic survey of DWMs according to their prediction modalities, summarizing the unique challenges and development paths for each modality. We also summarize the applications of DWMs and their comprehensive impact on AD research. Next, this paper compares the performance of representative works in major prediction and planning tasks. Finally, we discuss the limitations of current DWMs and identify promising research directions to overcome existing difficulties and promote continuous progress in this field. We believe that this survey will help new researchers to understand the key technical developments in this area quickly.
{
    \small
    \bibliographystyle{ieeenat_fullname}
    \bibliography{main}
}
\end{document}